\documentclass[letterpaper, 10 pt, conference]{ieeeconf}  % Comment this line out if you need a4paper

\IEEEoverridecommandlockouts                              % This command is only needed if 
\usepackage{graphics} % for pdf, bitmapped graphics files
\usepackage{epsfig} % for postscript graphics files
\usepackage{mathptmx} % assumes new font selection scheme installed
\usepackage{times} % assumes new font selection scheme installed
\usepackage{caption}

\usepackage[T1]{fontenc} % optional
\usepackage{cite}
\usepackage{booktabs}
\usepackage{amsmath,amssymb,amsfonts}
\usepackage{algorithmic}
\usepackage{setspace}
\usepackage{subfigure}
\usepackage{graphicx}
\usepackage{textcomp}
\usepackage[table]{xcolor}
\usepackage{multirow}
\usepackage{threeparttable}
\usepackage{tabularx}
\usepackage{mathtools}
\usepackage[free-standing-units=true]{siunitx}
\usepackage{float}
\newcommand{\rom}[1]{\uppercase\expandafter{\romannumeral #1\relax}}

\usepackage{array}
\newcommand{\PreserveBackslash}[1]{\let\temp=\\#1\let\\=\temp}
\newcolumntype{C}[1]{>{\PreserveBackslash\centering}p{#1}}
\newcolumntype{R}[1]{>{\PreserveBackslash\raggedleft}p{#1}}
\newcolumntype{L}[1]{>{\PreserveBackslash\raggedright}p{#1}}

\newcommand{\fref}[1]{Fig. \ref{#1}}
\newcommand{\sref}[1]{Section \ref{#1}}
\newcommand{\tref}[1]{Table \ref{#1}}
\newcommand{\aref}[1]{Appendix \ref{#1}}
\newcommand{\eref}[1]{Eq. \ref{#1}}

\newcommand{\monoicon}{\raisebox{-0.15\height}{\includegraphics[height=.9em]{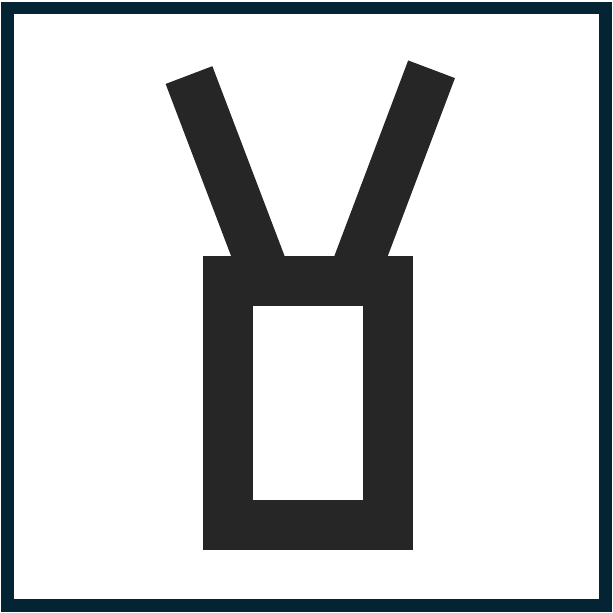}}}
\newcommand{\stereoicon}{\raisebox{-0.15\height}{\includegraphics[height=.9em]{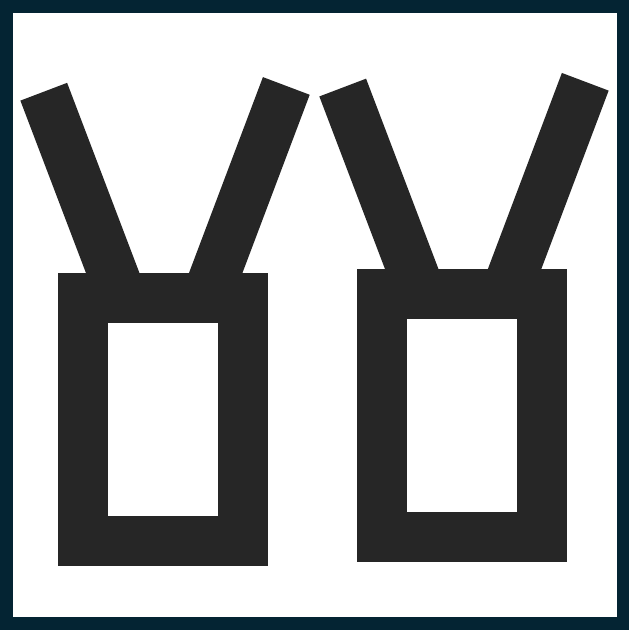}}}

\newcommand\Disparity{{\mathrm{Disp}}}

\DeclareMathOperator*{\argmin}{arg\,min}
\definecolor{ourgray}{gray}{0.9}

\makeatletter
\let\NAT@parse\undefined
\makeatother

\usepackage[colorlinks]{hyperref}
  \hypersetup{
    citecolor=blue,
    linkcolor=blue,   
    urlcolor=blue}

\author{Yuheng Qiu$^{*}$$^{1}$, Yutian Chen$^{*}$$^{1}$, Zihao Zhang$^{2}$, Wenshan Wang$^{1}$ and Sebastian Scherer$^{1}$% <-this % stops a space
\thanks{$^{*}$Equal contribution.}
\thanks{This work was supported by the DSTA under contract number \#DST000EC124000205.}
\thanks{$^{1}$Yuheng Qiu, Yutian Chen and Sebastian Scherer are with the Robotics Institute, Carnegie Mellon University, Pittsburgh, PA 15213, USA {\tt\small \{yuhengq, yutianch, basti\} @andrew.cmu.edu; }}
\thanks{$^{2}$ Zi hao Zhang is with the School of Ocean and Civil Engineering, Shanghai JiaoTong University, Shanghai 20024, China {\tt\small zihao6061@gmail.com}}%
}

\begin{document}

\title{\bf{\LARGE
MAC-VO: Metrics-aware Covariance for Learning-based \\Stereo Visual Odometry\\}
\bf\Large{\href{https://mac-vo.github.io}{mac-vo.github.io}}
}

\maketitle

\begin{abstract}
We propose MAC-VO, a novel learning-based stereo visual odometry (VO) framework that trains a metrics-aware uncertainty model to serve two critical functions: selecting keypoints and weighting residuals in pose graph optimization. 
Unlike traditional geometric methods that favor texture-rich features like edges, our keypoint selector leverages this learned uncertainty model to eliminate low-quality features based on global inconsistency. 
In contrast to learning-based approaches that rely on scale-agnostic weight matrices for covariance, our metrics-aware covariance model—derived from the learned uncertainty—captures spatial errors in keypoint registration and inter-axis correlations. 
By embedding this covariance model into pose graph optimization, MAC-VO achieves superior robustness and accuracy in pose estimation, excelling in challenging environments with varying illumination, feature density, and motion patterns. 
Evaluations on public benchmark datasets demonstrate that MAC-VO surpasses existing VO algorithms and even some SLAM systems in difficult scenarios. 
Additionally, the uncertainty map offers valuable insights for decision-making.
%\textbf{}
% The code and demo is available at\href{https://mac-vo.github.io}{https://mac-vo.github.io}
\end{abstract}


\section{Introduction}

\PARstart{V}{isual} Odometry (VO) predicts the relative camera pose from image sequences and often serves as the front-end of Simultaneous Localization and Mapping (SLAM) systems. 
Over the past few decades, both geometric and learning-based methods have been developed with significant advances in generalizability and accuracy~\cite{mur2015orb,tartanvo2020corl,teed2021droid,teed2024deep}.
However, VO remains a challenging problem in real-world scenarios, with multiple visually degraded scenarios such as low illumination, dynamic and texture-less scenes.

To improve the robustness in challenging scenes, geometric-based VO algorithms employ outlier filtering strategies \cite{fischler1981random} and weigh the optimization residuals by the covariance of the observed features \cite{zhao2021super}.
However, how to effectively \textit{select the reliable keypoints} and \textit{model their covariance} becomes two significant challenges. 
Existing methods typically select the keypoints based on local intensity gradient with a manually defined threshold \cite{superpoint2018, orb2011, sift2004}. 
These approaches lead to errors and outliers because they don't model the structure or context information of the environment (e.g., features on repetitive patterns may not be ideal candidates despite high image gradients). 
Moreover, the covariance model is often empirically modeled using a constant parameter, which is sub-optimal. 
In addition, the parameters in the keypoint selection and covariance model need to be extensively tuned for different environments. 

\begin{figure}[t]
    \captionsetup{font=small}
    \centering
    \includegraphics[width=.93\linewidth]{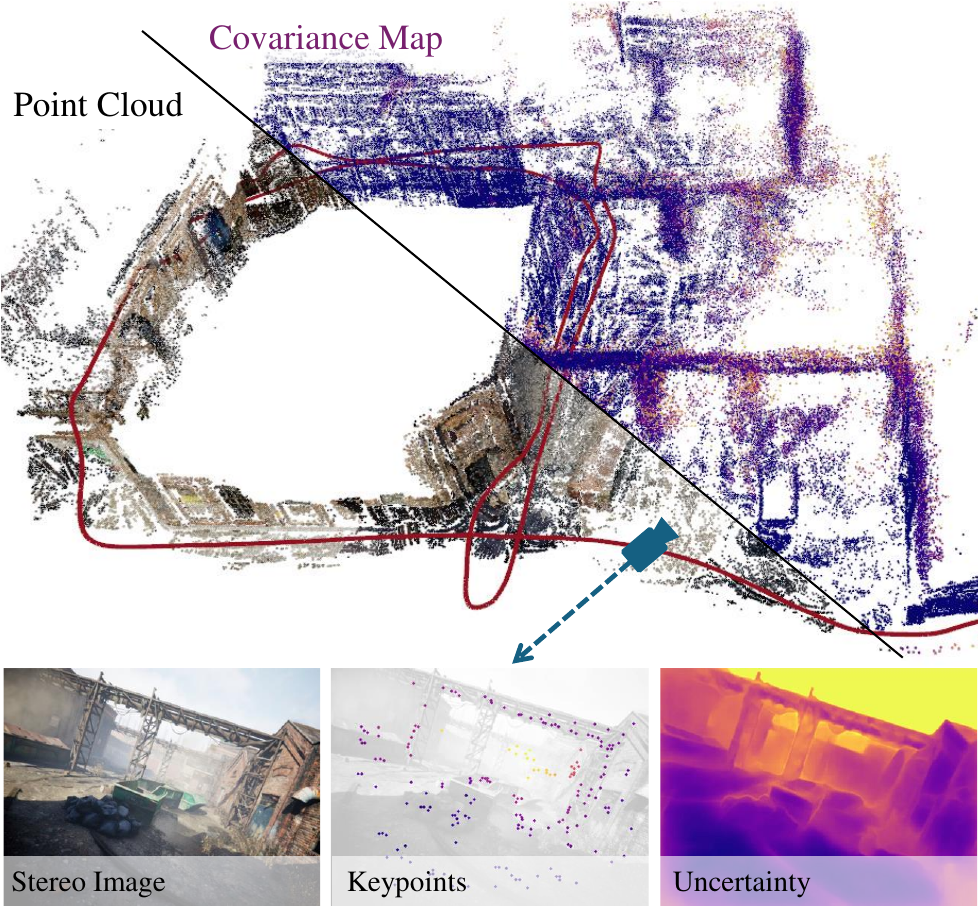}
    \captionsetup{width=\linewidth}
    \caption{\small Without relying on multi-frame bundle adjustment, {MAC-VO} aggregately reconstructs the map based on the two-frame Pose graph optimization. 
    We propose \textit{metrics-aware covariance} model for 3D keypoint based on the learning matching uncertainty.
    % The covariance model serves dual purposes: selecting keypoints and weighing the residual in the pose graph optimization. 
    }
    \label{fig:eyecather}
    \vspace{-10pt}
\end{figure}

With advances in learning-based visual features, more algorithms utilize learned features \cite{superpoint2018} to optimize the camera pose. Confidence score \cite{wang2023dust3r} or confidence weights \cite{teed2021droid, teed2024deep} of these feature points are often obtained in an unsupervised manner.
The learned confidence helps to track the features and model the reliability during the optimization. However, these confidence or uncertainty values are scale-agnostic, which means they don't reflect the actual estimation error in the 3D space. This scale-agnostic problem brings two limitations. 
Firstly, it makes the covariance inconsistent across different environments that vary in scale, such as indoors and outdoors. % Inter frame consistency
Secondly, it makes it harder to integrate multiple constraints from different modalities or sensors.

To overcome the above challenges, this paper addresses the problem of \textit{modeling metrics-aware covariance values for the 3D keypoints}. More specifically, this is achieved through two innovations.
% \wsnote{Is this the key point, or do we want to emphasize the modeling of metrics-aware uncertainty in 3D?} 
Firstly, we propose a learning-based model to quantify the 2D metrics-aware uncertainty of feature matching. 
Inspired by the FlowFormer \cite{huang2022flowformer} and GMA \cite{jiang2021learning}, we employ an iterative update model and motion aggregator to predict the uncertainty in 2D image space, which helps to filter unreliable features in the occluded region or low-illumination area.
Secondly, based on the learned 2D uncertainty values, we model the covariance of the feature points {in 3D space} using a \textit{metrics-aware 3D covariance model}.
Compared to DROID-SLAM~\cite{teed2021droid}, which utilizes a scale-agnostic diagonal covariance matrix, our approach provides a more accurate representation by modeling the covariance of 3D feature points. 
This covariance model includes the inter-axes correlation of the 3D features.
In the ablation study, we demonstrate the inter-frame consistency and the intra-frame consistency of our proposed covariance model.

We integrate the above two innovations into MAC-VO, a stereo VO that features superior keypoint selection and pose graph optimization based on the metrics-aware covariance model and achieves accurate tracking results in challenging cases compared with state-of-the-art VOs and even some SLAM systems without fine-tuning and without multi-frame optimization. 
% \yuheng{more about the system, like without multi-frame like accuracy}
In summary, the main contributions are: 

\begin{itemize}
    \itemsep0em
    \item 
    We present a learning-based 2D uncertainty network with metrics awareness, leveraging an iterative motion aggregator to capture the inconsistency of the feature matching. 
    We use this metrics-aware uncertainty to evaluate the quality of features, in order to select keypoints, and guide the backend optimization.

    \item This paper introduces a novel metrics-aware 3D covariance model based on the 2D uncertainty of feature matching and depth estimation. 
    The ablation study demonstrates the necessity of scale consistency and the off-diagonal terms in the pose graph optimization. 

    \item We propose the MAC-VO, a stereo VO pipeline that estimates the camera pose and registers 3D features with metrics-aware covariance. 
    In the experiments, MAC-VO outperforms existing VO algorithms even some SLAM algorithms in challenging environments. 
    
\end{itemize}

\begin{figure}
    \centering
    \captionsetup{font=small}
    \includegraphics[width=\linewidth]{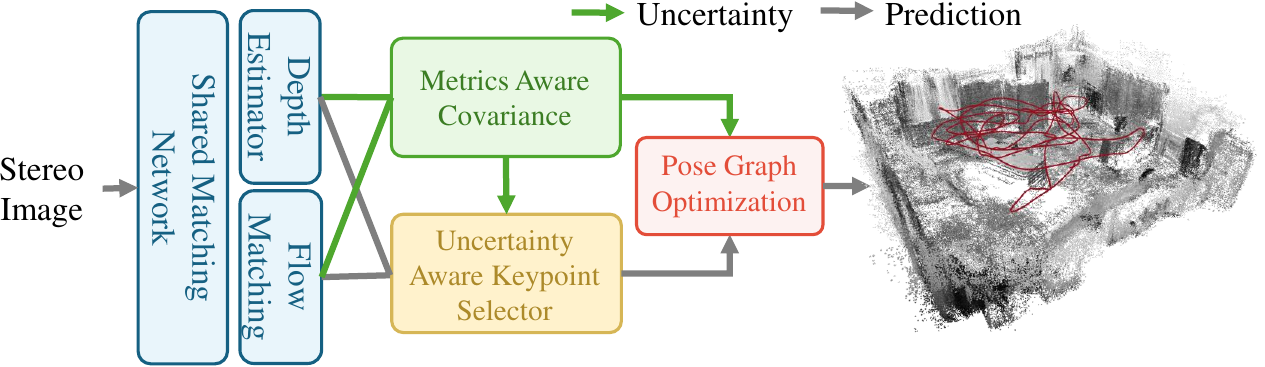}
    \caption{MAC-VO pipeline. It employs a shared matching network to jointly estimate depth, optical flow, and corresponding uncertainties. The learned uncertainty is utilized in the keypoint selector, the metrics-aware 3D covariance model, and the back-end optimization.}
    % \caption{MAC-VO pipeline. We use a shared matching network to estimate the depth, flow, and corresponding uncertainty. The learned uncertainty is leveraged to filter out unreliable features and model the metrics-aware covariance of the 3D keypoints. These registered keypoints and their covariance models are then utilized in the back-end optimization to determine the relative camera pose. }
    \label{fig: Pipeline}
    \vspace{-15pt}
\end{figure}

\section{Related Works}

Existing geometric-based methods optimize the camera pose based on geometric constraints like re-projection error \cite{mur2015orb, klein2007parallel} or photometric error \cite{wang2017stereo, engel2014lsd, gao2018ldso}.
To more accurately model the uncertainty of the depth, Civera et al. \cite{Javier2008InverseDepth} and Montiel\cite{Montiel2006UnifiedInverseDepth} investigate the inverse depth parameterization.
These methods often use constant parameters and simple heuristics to model the covariance matrix of these errors during the factor graph optimization.
For multi-sensor fusion \cite{zhao2021super, qin2018vins} and semantic SLAM \cite{yang2019cubeslam, qiu2022airdos}, the covariance model plays a significant role in weighting the confidence of different sensors and modules.
To capture sensor uncertainty, the covariance models are often tuned based on empirical prior. 
However, these simplified covariance models fail to capture the complexity of the challenging environments.

% \paragraph{Learning-based Methods}
Recent advances in deep learning have transformed research in optical flow estimation \cite{teed2020raft, parameshwara2022diffposenet}, feature matching \cite{sarlin2020superglue, wang2023dust3r}, depth estimation \cite{zhou2017unsupervised, li2018undeepvo, tateno2017cnn}, and end-to-end camera pose estimation \cite{costante2015exploring, tartanvo2020corl}.
Several methodologies have been developed to address the uncertainty in estimating depth, flow, and pose.
Dexheimer et al. \cite{Dexheimer_2023_learndepthcov} proposed a learned depth covariance function that is applied in downstream tasks like 3D reconstruction.
Nie et al. utilize a self-supervised learning method to jointly train depth and depth covariance of images in the wild \cite{Xinyu_2022_SelfImprovDepthEst}. 
ProbFlow \cite{Wannenwetsch_2017_ICCV} proposes using post-hoc confidence measures to assess the per-pixel reliability of flow.

Amidst these developments, hybrid learning-based SLAM systems \cite{yang2020d3vo, li2018undeepvo, fu2024islam, ranjan2019competitive, backtofeat2021, costante2020uncertainty} are emerging to synergize the geometric constraint with the adaptability of learning-based methods.
To improve the reliability of the feature tracking process, some methods introduce learning-based uncertainty measurements or confidence scores \cite{muhle2023learning, Kaygusuz_2021, wang2023dust3r, rabiee2020ivslam, murai2024mast3r} for pose optimization.
DROID-SLAM \cite{teed2021droid} utilizes a differentiable bundle adjustment layer to implicitly tune the uncertainty model. This method employs a simplified diagonal covariance model for the bundle adjustment.
These methods focus on the relative confidence between features, but ignore the scale consistency of the covariance model.

For keypoint selection, geometric-based VO relies on hand-crafted features \cite{orb2011, sift2004}, which detect the edges and corner points. 
The recent advance of the learning-based method train feature extractor in data-driven manner \cite{superpoint2018, pixelperfect2021}. 
However, these keypoint also prioritize the edge and corner features due to their data bias in the pre-training dataset.
Recent works like D3VO \cite{yang2020d3vo} have shown that relying on edge and corner points can degrade state estimation, sometimes performing worse than random selectors \cite{teed2024deep}.
The accuracy of learning-based feature matching and depth estimation algorithms is particularly compromised at object edges due to feature interpolation and the ambiguity in neural networks.
In this work, we propose a keypoint selector based on learned uncertainty to filter out the unreliable features.
% These methods, however, either adopt a simplified diagonal covariance model or are scale-agnostic, resulting in suboptimal performance in 

% \wsnote{why don't we mention that the uncertainty model of DROID-SLAM is insufficient? }

\section{Method}

\begin{figure*}[t]
    \centering
    \captionsetup{font=small}
    \includegraphics[width=\linewidth]{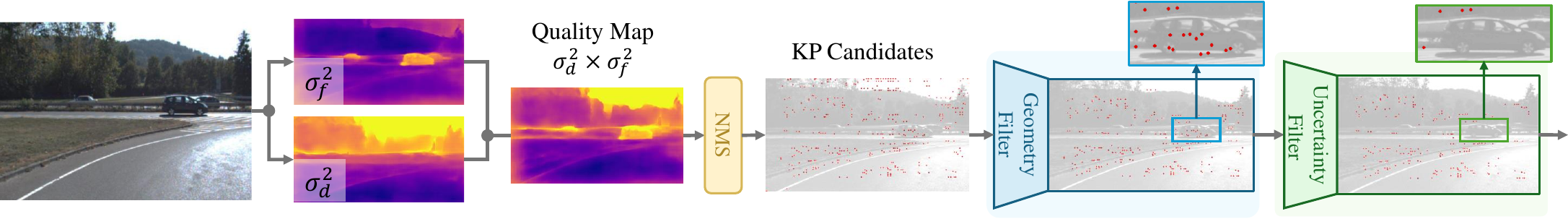}
    \caption{We include three filters: Non-minimum Suppression (NMS) filter, geometric filter, and uncertainty-based filter. In the KITTI dataset, the uncertainty filter implicitly filters out the unreliable feature matchings in the scene. }
    \label{fig:KeypointSelector}
    \vspace{-15pt}
\end{figure*}

As illustrated in \fref{fig: Pipeline}, MAC-VO outlines an effective integration of a learning-based front-end and a geometrically constrained back-end using the metrics-aware covariance model.
In the front-end (Section \ref{Method: Network}), we train an uncertainty-aware matching network to model the corresponding uncertainty stems from feature deficiencies. 
Utilizing the learned uncertainty, we develop a keypoint selector in Section \ref{Method: keypointslection} to choose reliable features.
% In the back-end optimization \wsnote{why do we talk about D before C?}(Section \ref{Method: Optimization}), we optimize the relative motion by minimizing the distance between the registered keypoints.
% To bridge the gap between the learning-based front-end and the geometry-based back-end optimization, we propose a metrics-aware 3D covariance model in Section \ref{Method: covmodel}, which projects the 2D uncertainty into 3D space.
{The metrics-aware 2D uncertainty is then propagated to the 3D space via the proposed covariance model in \sref{Method: covmodel}. In the back-end optimization (\sref{Method: Optimization}), we optimize the relative motion by minimizing the distance between registered keypoints weighted by the 3D covariance.}

\begin{figure}
    \centering
    \captionsetup{font=small}
    \includegraphics[width=\linewidth]{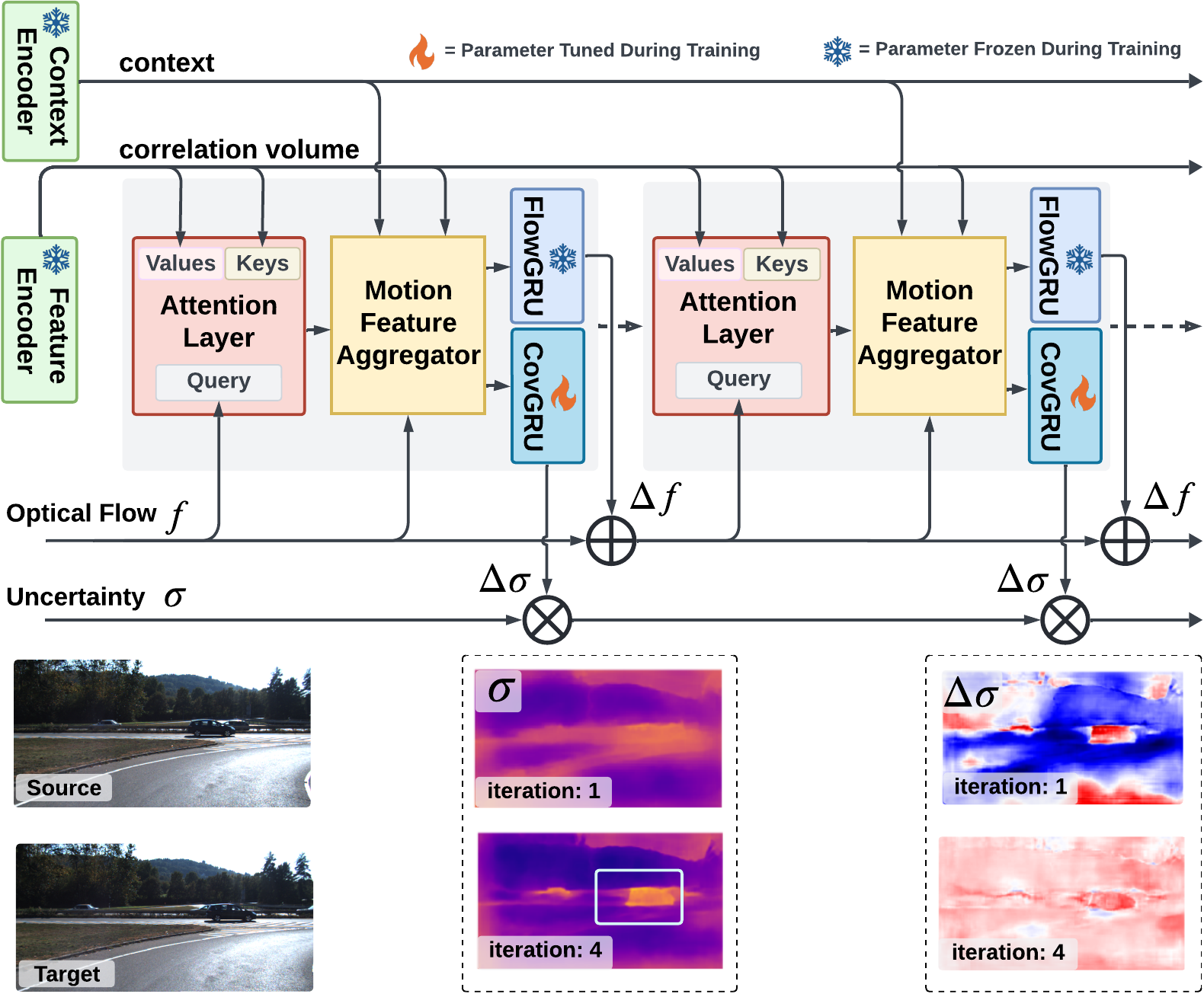}
    \caption{\textbf{Top: } Architecture of the uncertainty-aware matching network. 
    % We employ a motion aggregator and an iterative update structure to enable the covariance module to capture inconsistencies in feature matching.
    \textbf{Bottom: } Each iteration, the model captures the inconsistency of the matching. For the $\Delta \sigma$, the {\color{red}red} color indicates a positive $\Delta \sigma$ increasing the uncertainty, and {\color{blue}blue} means decreasing uncertainty.
    }
    \label{fig: FlowFormerCov}
    \vspace{-10pt}
\end{figure}

\subsection{Network \& Uncertainty Training}
\label{Method: Network}

The objective of the proposed matching network is to estimate the flow $\hat{f} \in \mathbb{R}^2$ and the corresponding uncertainty $\hat\Sigma_f = \mathrm{diag}(\sigma_u^2, \sigma_v^2)$ between two frames. 
We adopt the FlowFormer architecture \cite{huang2022flowformer} as the backbone for optical flow estimation, leveraging its transformer-based design.
As shown in \fref{fig: FlowFormerCov}, the recurrent decoder iteratively refines flow predictions by capturing ambiguities in feature correspondence within the cost volume’s matching space, utilizing global context from the encoded cost memory.

To extend this framework and leverages both global motion cues and local feature, we introduce a covariance decoder that predicts $\log \Delta \sigma$, the iterative updates of the uncertainty on log space.
Operating the uncertainty update on the log space facilitates additive updates, ensures positive variances, and stabilizes gradients.
After iterative updates, the log-variance passes through an \texttt{exp} activation function to obtain the final uncertainty. More details about the network and training are shown in \aref{appendix:Network}.

To supervise the uncertainty, we leverage the negative log-likelihood loss used in conformal prediction \cite{russell2021multivariate, ConformalPrediction, qiu2023airimu}:
\begin{equation}
     L_{cov} = \sum_i^N \alpha_i \left((y - \hat{f}_i)^\top \hat{\Sigma}_{f}^{-1} (y - \hat{f}_i) + \log (\det \hat{\Sigma}_f) \right),
\end{equation}
where $y$ is the ground truth optical flow, $f_i$ denotes the $i$-th iteration of the network outputs, and $\alpha_i$ is the weight for each iteration and is set to decrease exponentially with a ratio of $0.8$.
During the training stage, we initialize the encoder network parameters with the pre-trained model by FlowFormer.
We then train the covariance module on the synthetic dataset TartanAir \cite{tartanair2020iros}, and evaluate the model in the TartanAirv2 sequences and other real-world datasets.
Our experiments demonstrate that the model can generalize to real-world datasets without fine-tuning.

\subsection{Uncertainty-based Keypoint Selection}
\label{Method: keypointslection}

Different from the random selector used in DPVO\cite{teed2024deep} and the hand-crafted features selector used in ORB-SLAM \cite{orb2011}, we leverage the learned uncertainty estimation to filter out unreliable features. 
This is achieved by composing three filters: the \textit{uncertainty filter}, \textit{geometry filter}, and the \textit{non-minimum suppression} (NMS).
To prevent the clustering of keypoint candidates, we first apply the NMS filter, ensuring a more even spatial distribution of features.
The geometry filter further removes keypoints located at the image boundaries and those outside the valid depth observation range.
Finally, the uncertainty filter eliminates pixels with depth and flow uncertainty greater than 1.5 times the median uncertainty of the current frame.
Illustrated in \fref{fig:KeypointSelector}, the uncertainty filter removes all keypoint candidates on the moving vehicle on a KITTI trajectory. 
Our proposed uncertainty estimation strategy also filters out keypoints on occluded objects, reflective surfaces, and feature-less areas, resulting in a more robust and reliable feature set.
% Along with the uncertainty filter is the geometric filter, which constrains removes keypoints on the edge of the frame. 

% After this preliminary filtering, the remaining points will pass through two filters:  \textit{uncertainty filter} and \textit{geometry filter}. 
% The uncertainty filter removes the points with high depth and flow uncertainty in the upper quartile of all candidate points. It implicitly filters out unreliable keypoints on occluded objects, reflective surfaces, and feature-less areas as shown in \fref{fig:KeypointSelector}.
% \yuheng{remove a little bit about geometry and add more about the figure}
% Geometry filters, similar to those in traditional SLAM algorithms, employ two key rules: (1) the maximum depth of keypoints should be within a threshold and (2) the points should not be near the edge of the image. This filter ensures the minimum disparity of keypoints to be at least $1$ pixel and filters out unreliable points due to physical limitations.
% To make features evenly distributed in the image, we employ a non-\textit{minimum} suppression (NMS) filter on the product of flow and depth uncertainty
% In conclusion, the uncertainty-aware keypoint selector selects reliable keypoints with even spatial distribution.

\begin{figure}[t]
    \centering
    \captionsetup{font=small}
    \includegraphics[width=.9\linewidth]{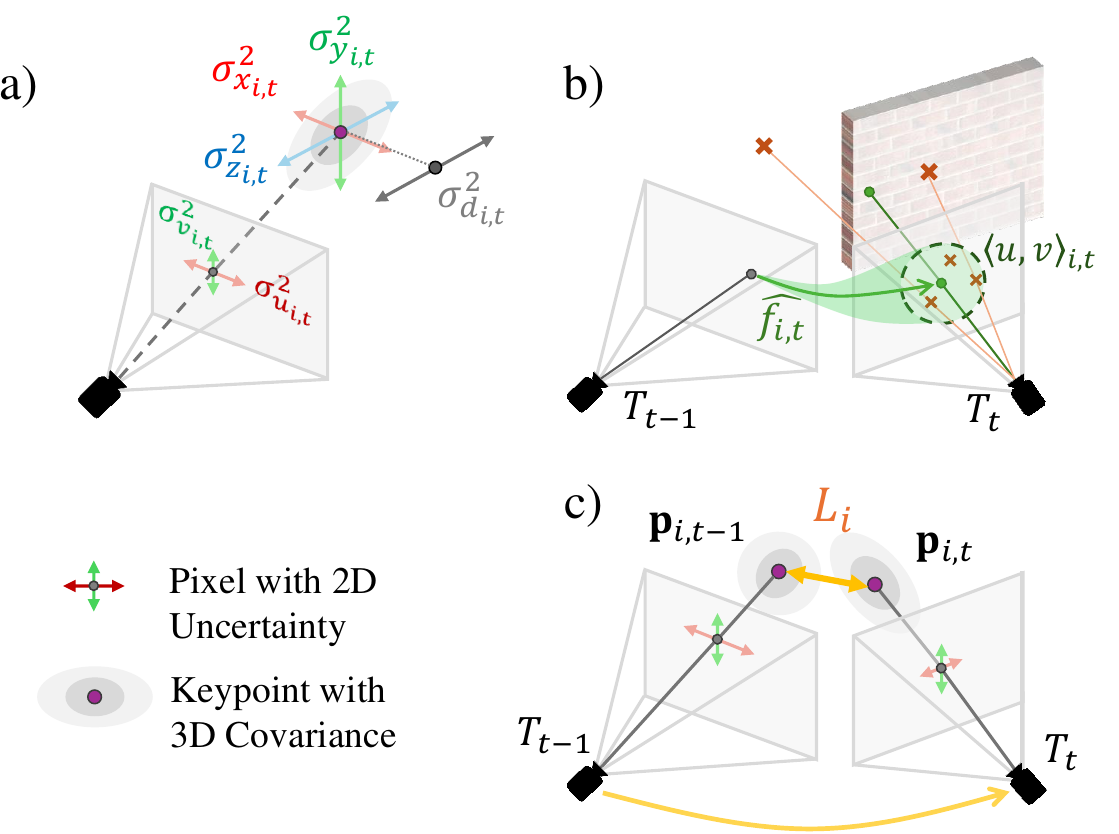}
    \caption{\textbf{a)} Uncertain estimation of depth due to the error in feature matching. \textbf{b)} Projecting depth and matching uncertainty from image plane to 3D space. \textbf{c)} Residual $\mathcal{L}_i$ for pose graph optimization.}
    \label{fig:Covariance}
    \vspace{-10pt}
\end{figure}

% In previous sections, we introduced the keypoint selector that benefits from the estimated covariance. 
% This section will detail how COVO incorporates the estimated covariance into pose graph optimization problem, guiding the Levenberg-Marquardt optimizer towards more accurate pose estimation.

% First, we will formulate the optimization problem and residue. We will then show how to project uncertainty on a 2D image plane to 3D space under world coordinates. Finally, we derive the covariance of residual and explain how optimizer leverages the covariance.
% In previous sections, we estimate the per-pixel depth uncertainty using estimated disparity uncertainty. This section, we formulate a method to estimate uncertainty with the presence of matching uncertainty.

\subsection{Metrics-aware 3D Covariance Model}
\label{Method: covmodel}
In the context of the camera projection geometry, the covariance of a 3D keypoint is determined by the uncertainty of the depth $\sigma_d^2$ and matching $(\sigma_u^2, \sigma_v^2)$. 
% \yuheng{add the symbol of the matched depth}
To accurately model the covariance for 3D keypoint, it is critical to determine (1) the depth uncertainty of the matched points $\hat{\sigma}_d^2$ and (2) the off-diagonal covariance terms during the 2D-3D projection. 

\textbf{Project 2D Uncertainty to 3D Covariance}\;
Following the pinhole camera model with focal length $\mathrm{f}_x, \mathrm{f}_y$ and optical center $\mathrm{c}_x, \mathrm{c}_y$, the coordinate of the keypoint is calculated by: $\mathbf{x}_{i, t} = (\mathbf{u}_{i, t} - \mathrm{c}_x)\mathbf{d}_{i, t} / \mathrm{f}_x$, $\mathbf{y}_{i, t} = (\mathbf{v}_{i, t} - \mathrm{c}_y)\mathbf{d}_{i, t} / \mathrm{f}_y$ and $\mathbf{z}_{i, t} = \mathbf{d}_{i, t}$, as shown in  \fref{fig:Covariance} (a). To accurately capture the uncertainties associated with these measurements, the main diagonal of the covariance matrix is formulated as:
\begin{equation}
\label{eq: Variance terms}
\begin{aligned}
    \sigma_{x_{i, t}}^2 &= {\left(\sigma_{u_{i, t}}^2\sigma_{d_{i, t}}^2 + \sigma_{u_{i, t}}^2d_{i, t}^2 + (u_{i, t} - c_x)^2\sigma_{d_{i, t}}^2\right)}/{\mathrm{f}_x^2},\\
    \sigma_{y_{i, t}}^2 &= {\left(\sigma_{v_{i, t}}^2\sigma_{d_{i, t}}^2 + \sigma_{v_{i, t}}^2d_{i, t}^2 + (v_{i, t} - c_y)^2\sigma_{d_{i, t}}^2\right)}/{\mathrm{f}_y^2},\\
    \quad\sigma_{z_{i, t}}^2 &= \sigma_{d_{i, t}}^2.
\end{aligned}
\end{equation}

In this model, the projected coordinates are interdependent due to the common multiplier of depth $\mathbf{d}_{i,t}$.
To precisely formulate the covariance of the 3D keypoints ${}^c\Sigma_{i, t}^p$ under camera coordinate, it is essential to include the off-diagonal covariance terms in ${}^c\Sigma_{i, t}^p$.  
\begin{equation}
    \label{eq: Covariance terms}
    \begin{aligned}
        {}^c\Sigma_{i, t}^p = \begin{bmatrix}
            \sigma_{z}^2 & \sigma_{xz_{i, t}} & \sigma_{yz_{i, t}} \\
            \sigma_{xz_{i, t}} & \sigma_{x_{i, t}}^2 & \sigma_{xy_{i, t}} \\
            \sigma_{yz_{i, t}} & \sigma_{xy_{i, t}} & \sigma_{y_{i, t}}^2 \\
        \end{bmatrix}\begin{aligned}
            &\sigma_{xz} = \sigma_{d}^2 (u - \mathrm{c}_x)/\mathrm{f}_x\\
            &\sigma_{yz} = \sigma_{d}^2 (v - \mathrm{c}_y)/\mathrm{f}_y\\
            &\sigma_{xy} = \frac{\sigma_{d}^2 (u - \mathrm{c}_x) (v - \mathrm{c}_y)}{\mathrm{f}_x\mathrm{f}_y}.
        \end{aligned}\\
    \end{aligned}
\end{equation}
Results in the ablation study also confirm the necessity of off-diagonal terms for accurate pose graph optimization. Detailed derivation for \eref{eq: Variance terms} and \eref{eq: Covariance terms} is in \aref{appendix:Project2DTo3D}. 

\textbf{Uncertainty Correction after Keypoint Matching}\;
As shown in \fref{fig:Covariance} (b), the matched features are expected to fall in a probabilistic range centered at $[u_{i, t}, v_{i, t}]$ with the flow uncertainty $(\sigma_{u_{i, t}}^2, \sigma_{v_{i, t}}^2)$. However, the depth observations of matched keypoints may vary due to scene geometry.
For example, if a matched keypoint is located near the edge of an object, such as a wall, even a minor perturbation in feature matching can lead to substantial discrepancies in 3D keypoint registration. This, in turn, amplifies the uncertainty in the reconstructed 3D structure.

To address this problem, we correct the depth uncertainty $\hat{\sigma}_{d_{i, t}}^2$ of the matched feature point based on the depth feature of the local patch $D_{i,t}$ with the kernel size of $32$. 
We approximate it with the weighted sum of the variances within the patch. The weights are determined by a 2D Gaussian kernel $\varphi$, which utilizes $\sigma_{u_{i, t}}^2$ and $\sigma_{v_{i, t}}^2$ to adjust the influence of each point within the patch:
$\sigma_{d_{i, t}}^2 = \sum_j{\varphi_j (d_j - \mu_{{D}_{i, t}})^2}.$ 
\vspace{4pt}

\subsection{Pose Graph Optimization}
\label{Method: Optimization}

We optimize the camera pose  $T_{t}\in SE(3)$ at time $t$ in the world frame by minimizing the distance of the matched 3D keypoints $p_{i, t-1}$ and ${}^cp_{i, t}$, where ${}^cp_{i, t}$ is in the camera frame.
To reduce the initial error margin of the optimization, we initialize the camera pose using the relative motion estimated by the TartanVO \cite{tartanvo2020corl}.

The pose graph optimization is formulated as follows:
\begin{equation}
\begin{aligned}
    T^\star &= \argmin_{T_t}{\sum_{i}{\left\| \mathbf{p}_{i, t-1} - {T}_{t}\, {}^c\mathbf{p}_{i, t} \right\|^2_{\Sigma_i}}}, \\
    \Sigma_{i} &= \Sigma_{i, t-1}^p + R_t\;{}^c\Sigma_{i, t}^p\;R_t^\top.
\end{aligned}
\end{equation}
$\|\cdot\|_{\Sigma_i}$ represents the Mahalanobis distance with covariance matrix $\Sigma_i$.
Unlike DROID-SLAM \cite{teed2021droid}, which employs a diagonal covariance matrix, we model the correlation between axes to capture accurate inter-dependencies.
We solved this problem by Levenberg-Marquardt using \textit{PyPose} \cite{wang2023pypose, zhan2023pypose}.

\section{Experiment}

\textbf{Datasets \& Baseline}\; We evaluate the proposed model and baseline methods on a variety of public datasets, including synthetic dataset TartanAir v2 \cite{tartanair2020iros}, real-world data from EuRoC \cite{euroc2016}, KITTI \cite{kitti2012}, as well as customized data collected from a Zed-X
% \footnote{\href{https://www.stereolabs.com/products/zed-x}{https://www.stereolabs.com/products/zed-x}} 
camera. 
These datasets cover a diverse range of hardware configurations, motion patterns, and environments. 
To demonstrate our method's robustness under challenging scenarios, we collected the TartanAir v2, a new set of difficult trajectories following the TartanAir \cite{tartanair2020iros} that includes frequent indoor-outdoor transition and low-illumination scenes as shown in \fref{fig:qualitative}. 
To demonstrate the generalizability of our model, we use the same configuration and same matching model across all datasets.

\textbf{Evaluation Metrics}\; Since the proposed method does not contain loop closure or global bundle adjustment, our evaluation focuses on relative error.
We use relative translation error ($t_{\mathrm{rel}}$, $\mathrm{m/frame}$) and rotation error ($r_{\mathrm{rel}}$, $\mathrm{{}^\circ/frame}$) as:
% \begin{equation}
%     \resizebox{.93\linewidth}{!}{%
%     $t_{\mathrm{rel}} = \frac{1}{N}\sum^N_{t=1}{\left\| p_{t+1} - p_{t} - R_{t}\hat{R}_{t}^\top\!\!\left(\hat{p}_{t+1} - \hat{p_{t}}\right)\right\|_2},\; 
%     r_{\mathrm{rel}} = \frac{180}{\pi}\frac{1}{N} \sum^N_{t=1}\left\| \log\left( \hat{R}_{t,t+1}^\top  R_{t,t+1}\right)\right\|_2, $}
% \end{equation}
\begin{equation}
\begin{aligned}        
    t_{\mathrm{rel}} &= \frac{1}{N}\sum^N_{t=1}{\left\| \mathbf{p}_{t+1} - \mathbf{p}_{t} - R_{t}\hat{R}_{t}^\top\!\!\left(\hat{\mathbf{p}}_{t+1} - \hat{\mathbf{p}_{t}}\right)\right\|_2},\\
    r_{\mathrm{rel}} &= \frac{180}{\pi}\frac{1}{N} \sum^N_{t=1}\left\| \log\left( \hat{R}_{t,t+1}^\top  R_{t,t+1}\right)\right\|_2, 
\end{aligned}
\end{equation}

where $\mathbf{p}_t$ and $R_t$ are ground truth position and rotation, $\hat{\mathbf{p}}_t$ and $\hat{R}_t$ is the estimated position and rotation. $R_{t,t+1}= R_{t}^\top R_{t+1}$ is the rotation from frame $t$ to frame $t+1$.

\subsection{Quantitative Analysis}
\begin{table*}[htbp]
\centering
\captionsetup{font=small}
\caption{Performance comparison of different methods on the EuRoC Dataset. Only odd-ordered trajectory is shown due to page limit, see \aref{appendix:AdditionalResults} for the remaining results. Average is calculated over all trajectories of EuRoC.}
\vspace{-2mm}
\resizebox{\linewidth}{!}{
\begin{tabular}{lcccccccccccccc}
\toprule
 Trajectory & \multicolumn{2}{c}{{MH01}} & \multicolumn{2}{c}{{MH03}} & \multicolumn{2}{c}{{MH05}} & \multicolumn{2}{c}{V102} & \multicolumn{2}{c}{V201} & \multicolumn{2}{c}{V203} & \multicolumn{2}{c}{Avg.} \\
\cmidrule{2-15}
   &  $t_{\mathrm{rel}}$ & $r_{\mathrm{rel}}$ & $t_{\mathrm{rel}}$ & $r_{\mathrm{rel}}$ & $t_{\mathrm{rel}}$ & $r_{\mathrm{rel}}$ & $t_{\mathrm{rel}}$ & $r_{\mathrm{rel}}$ & $t_{\mathrm{rel}}$ & $r_{\mathrm{rel}}$ & $t_{\mathrm{rel}}$ & $r_{\mathrm{rel}}$ & $t_{\mathrm{rel}}$ & $r_{\mathrm{rel}}$ \\
\midrule
\textbf{SLAM} \\
\hspace{3mm} \stereoicon~ORB-SLAM 3                 & 0.0035   & 0.0450 & 0.0058   & 0.0603 & 0.0059   & 0.0526 & 0.0096   & 0.1757 & 0.0064   & 0.1615 & 0.033    & 0.9497 & 0.0092 & 0.1866 \\
\hspace{3mm} \stereoicon~DROID-SLAM$^{\star}$          & \textbf{0.0012} & \textbf{0.0159} & 0.0034   & 0.2656 & \textbf{0.0025} & \textbf{0.0193} & \textbf{0.0026} & \textbf{0.0417} & \textbf{0.0012} & \textbf{0.0289} & \textbf{0.0034} & \textbf{0.1033} & \textbf{0.0024} & 0.0590 \\
\hspace{3mm} \monoicon~MASt3R-SLAM$^{\star}$ & 0.0143 & 0.5493 & 0.0493 & 0.8890 & 0.0254 & 0.5808 & 0.0354 & 1.3178 & 0.0117 & 0.6818 & 0.0430 & 2.3306 & 0.0277 & 1.0472 \\
\midrule
\textbf{VO} \\
\hspace{3mm} \monoicon~TartanVO$^{\star}$   & 0.0121 & 0.0560 & 0.0302 & 0.2791 & 0.0193 & 0.0604 & 0.0251 & 0.1244 & 0.0065 & 0.0920 & 0.0303 & 0.2986 & 0.0198 & 0.1270 \\
\hspace{3mm} \stereoicon~TartanVO                   & 0.0277 & 0.5122 & 0.0514 & 0.6635 & 0.0464 & 0.4797 & 0.0394 & 1.0420 & 0.0195 & 0.4684 & 0.0473 & 1.9657 & 0.0368 & 0.8346 \\
\hspace{3mm} \stereoicon~iSLAM-VO         & 0.0042 & 0.0560 & 0.0076 & 0.2789 & 0.0070 & 0.0603 & 0.0066 & 0.1241 & 0.004  & 0.0920 & 0.0151 & 0.2984 & 0.0071 & 0.1269 \\
\hspace{3mm} \monoicon~DPVO$^{\star}$      & 0.0015 & \underline{0.0207} & \underline{0.0028} & \underline{0.0273} & \underline{0.0028} & 0.0243 & 0.0041   & 0.0496 & \underline{0.0016} & \underline{0.0342} & \underline{0.0045} & \underline{0.1205} & \underline{0.0027} & \underline{0.0437} \\
% \midrule
\rowcolor{ourgray}
\hspace{3mm} \stereoicon~\textbf{Ours}                  & \underline{0.0014} & 0.0214 & \textbf{0.0023} & \textbf{0.0238} & \textbf{0.0025} & \underline{0.0216} & \underline{0.0029} & \underline{0.0434} & \textbf{0.0012} & \textbf{0.0289} & 0.0049  & 0.1284 & \textbf{0.0024} & \textbf{0.0403} \\
\bottomrule
\multicolumn{12}{l}{
% $^\ddagger$ Average over all trajectories of EuRoC.
\quad
\monoicon~\hspace{.5mm} Monocular method.
\quad
\stereoicon~\hspace{.5mm} Stereo method.
\quad
$^\star$ Scale-aligned with ground truth.} \\
\end{tabular}
}
\vspace{-5pt}
\label{tab:EuRoCOdd}
\end{table*}

\textbf{EuRoC Dataset} \;\;
We assessed our model on the EuRoC \cite{euroc2016} dataset, as detailed in \tref{tab:EuRoCOdd}, comparing it against baseline methods including visual odometry and state-of-the-art visual SLAM systems with loop-closure and global bundle adjustment. While our method exhibits compatible performance to DROID-SLAM on average $t_{\mathrm{rel}}$, it outperforms all baselines in terms of $r_{\mathrm{rel}}$ by around $10\%$. 

% \begin{figure}
%     \centering
%     \captionsetup{font=small}
%     \includegraphics[width=\linewidth]{figs/TartanAir2.pdf}
%     \caption{TartanAir v2, a dataset for visual SLAM with abrupt changes in illumination and aggressive motion.}
%     \label{fig:tartanair-v2}
%     \vspace{-5pt}
% \end{figure}

\begin{table*}[htbp]
\centering
\captionsetup{font=small}
\caption{Performance comparison on the TartanAir v2 Hard Dataset. Noticeably, ORB-SLAM3 and MASt3R-SLAM lost track on all sequences and are not included. See \aref{appendix:AdditionalResults} for the \textit{Easy} subset. Average is calculated over all trajectories of TartanAir v2 Hard.}
% \vspace{-2mm}
\resizebox{\linewidth}{!}{
\begin{tabular}{lcccccccccccccccc}
\toprule
 Trajectory & \multicolumn{2}{c}{{H00}} & \multicolumn{2}{c}{{H01}}  & \multicolumn{2}{c}{{H02}} & \multicolumn{2}{c}{{H03}}  & \multicolumn{2}{c}{{H04}} & \multicolumn{2}{c}{{H05}}  & \multicolumn{2}{c}{{H06}} & \multicolumn{2}{c}{Avg.} \\
\cmidrule{2-17}
   &  $t_{\mathrm{rel}}$ & $r_{\mathrm{rel}}$ & $t_{\mathrm{rel}}$ & $r_{\mathrm{rel}}$ & $t_{\mathrm{rel}}$ & $r_{\mathrm{rel}}$ & $t_{\mathrm{rel}}$ & $r_{\mathrm{rel}}$ & $t_{\mathrm{rel}}$ & $r_{\mathrm{rel}}$ & $t_{\mathrm{rel}}$ & $r_{\mathrm{rel}}$ & $t_{\mathrm{rel}}$ & $r_{\mathrm{rel}}$ & $t_{\mathrm{rel}}$ & $r_{\mathrm{rel}}$ \\
\midrule
\textbf{SLAM} \\
% \hspace{3mm} \monoicon~MASt3R-SLAM$^{\star}$ & .4232 & 7.424 & .4844 & 7.280 & - & - & - & - & .3101 & 7.214 & .3297 & 7.589 & .3465 & 7.259 & .3788 & 7.353 \\
\hspace{3mm} \stereoicon~DROID-SLAM$^\star$           & \underline{.0485} & \underline{.1174} & \textbf{.0023} & \textbf{.0210} & \underline{.0190} & \underline{.0821} & \textbf{.0064} & \underline{.0300} & \textbf{.0057} & \textbf{.0255} & .1463 & \underline{.2357} & \underline{.0310} & \textbf{.0908} & \underline{.0370} & \textbf{.0861} \\
\midrule
\textbf{VO} \\
\hspace{3mm} \monoicon~TartanVO$^{\star}$  & .1605 & 3.338 & .2918 & 2.775 & .2718 & 3.305 & .2775 & 2.191 & .2204 & 2.874 & .1644 & 2.899 & .2350 & 3.756 & .2316 & 3.020 \\
\hspace{3mm} \stereoicon~TartanVO-stereo               & .1505 & .4329 & .0914 & .7542 & .0715 & .3265 & .0842 & .4053 & .0678 & .6569 & \underline{.0803} & 1.186 & .0784 & .9458 & .0892 & .6725 \\
\hspace{3mm} \stereoicon~iSLAM-VO         & .4235 & 2.630 & .3070 & 3.018 & .3252 & 2.189 & .3622 & 2.435 & .2576 & 2.899 & .2574 & 3.755 & .2099 & 3.145 & .3061 & 2.867 \\
\hspace{3mm} \monoicon~DPVO$^{\star}$       & .4984 & .4937 & .1738 & .7112 & .0539 & .2539 & .3847 & 2.703 & .0481 & .1869 & .1891 & 1.430 & .3365 & 2.943 & .2406 & 1.246 \\
% \midrule
\rowcolor{ourgray}
\hspace{3mm} \stereoicon~\textbf{Ours}                     & \textbf{.0085} & \textbf{.1018} & \underline{.0344} & \underline{.1450} & \textbf{.0048} & \textbf{.0628} & \underline{.0150} & \underline{.0778} & \underline{.0092} & \underline{.1414} & \textbf{.0048} & \textbf{.0552} & \textbf{.0217} & \underline{.4160} & \textbf{.0141} & \underline{.1429} \\
\bottomrule
\multicolumn{17}{l}{
\monoicon \hspace{.5mm} Monocular method.
\quad
\stereoicon \hspace{.5mm} Stereo method.
\quad
$^\star$ Scale-aligned with ground truth.} \\
\end{tabular}
}
\vspace{-10pt}
\label{tab: TartanAirHard Odd}
\end{table*}

\begin{table*}[htbp]
\centering
\captionsetup{font=small}
\caption{Performance comparison of different methods on the KITTI Dataset. Only even numbered trajectory is shown due to page limit, see \aref{appendix:AdditionalResults} for remaining results. Average is calculated over all trajectories of KITTI Odometry.}
% \vspace{-2mm}
\resizebox{\linewidth}{!}{
\begin{tabular}{lcccccccccccccc}
\toprule
 Trajectory & \multicolumn{2}{c}{{00}} & \multicolumn{2}{c}{{02}} & \multicolumn{2}{c}{{04}} & \multicolumn{2}{c}{06} & \multicolumn{2}{c}{08} & \multicolumn{2}{c}{10} & \multicolumn{2}{c}{Avg.} \\
\cmidrule{2-15}
   &  $t_{\mathrm{rel}}$ & $r_{\mathrm{rel}}$ & $t_{\mathrm{rel}}$ & $r_{\mathrm{rel}}$ & $t_{\mathrm{rel}}$ & $r_{\mathrm{rel}}$ & $t_{\mathrm{rel}}$ & $r_{\mathrm{rel}}$ & $t_{\mathrm{rel}}$ & $r_{\mathrm{rel}}$ & $t_{\mathrm{rel}}$ & $r_{\mathrm{rel}}$ & $t_{\mathrm{rel}}$ & $r_{\mathrm{rel}}$ \\
\midrule
\textbf{SLAM} \\
\hspace{3mm} \stereoicon~ORB-SLAM 3         & 0.0252   & 0.0586 & 0.0438   & 0.0529 & 0.0274   & 0.0322 & 0.0228   & 0.0338 & \underline{0.0271} & 0.0455 & \textbf{0.0166} & \underline{0.0515} & \textbf{0.0258} & \underline{0.0434} \\
\hspace{3mm} \stereoicon~DROID-SLAM$^\star$  & \underline{0.0198} & \underline{0.0538} & \underline{0.0250} & \underline{0.0445} & \underline{0.0255} & \underline{0.0296} & \underline{0.0199} & \underline{0.0296} & 0.0275   & \underline{0.0381} & 0.0309  & 0.0715 & 0.0900 & 0.0448 \\
\hspace{3mm} \monoicon~MASt3R-SLAM$^{\star}$ & 0.8988 & 0.7154 & 1.0104 & 0.5959 & 1.2266 & 0.1124 & 1.2599 & 0.1289 & 0.7760 & 0.6075 & 0.6889 & 0.4709 & 0.9538 & 0.4202 \\
\midrule
\textbf{VO} \\
\hspace{3mm} \monoicon~TartanVO$^{\star}$   & 0.2066   & 0.1055 & 0.1626   & 0.1105 & 0.1152   & 0.0789 & 0.2234   & 0.0816 & 0.1857   & 0.0823 & 0.1745  & 0.0907 & 0.2207 & 0.0886 \\
\hspace{3mm} \stereoicon~TartanVO                  & 0.0656   & 0.1026 & 0.0905   & 0.1197 & 0.1747   & 0.1158 & 0.0923   & 0.0968 & 0.0721   & 0.1063 & 0.0679  & 0.0969 & 0.1804 & 0.1147 \\
\hspace{3mm} \stereoicon~iSLAM-VO                   & 0.0577   & 0.1052 & 0.0686   & 0.1101 & 0.1356   & 0.0787 & 0.0837   & 0.0812 & 0.0510   & 0.082  & 0.0449  & 0.0905 & 0.0878 & 0.0883 \\
\hspace{3mm} \monoicon~DPVO$^{\star}$      & 0.4542   & \textbf{0.0495} & 0.4209   & \textbf{0.0381} & 0.0348   & \textbf{0.0219} & 0.2393   & \textbf{0.0250} & 0.3051   & \textbf{0.0347} & 0.0661  & \textbf{0.0386} & 0.1951 & \textbf{0.0329} \\
% \midrule
\rowcolor{ourgray}
\hspace{3mm} \stereoicon~\textbf{Ours}                           & \textbf{0.0192} & 0.0654 & \textbf{0.0223} & 0.0715 & \textbf{0.0206} & 0.0473 & \textbf{0.0187} & 0.0456 & \textbf{0.0254} & 0.0509 & \underline{0.019} & 0.0569 & \underline{0.0420} & 0.0645 \\
\bottomrule
\multicolumn{15}{l}{
% $^\ddagger$ Average over all trajectories (from 00 to 10) of KITTI Odom.
% \quad
\monoicon \hspace{.5mm} Monocular method.
\quad
\stereoicon \hspace{.5mm} Stereo method.
\quad
$^\star$ Scale-aligned with ground truth.} \\
\end{tabular}
}
\vspace{-5pt}
\label{tab:KITTIEven}
\end{table*}

\textbf{TartanAir v2 Dataset} \;\;
TartanAir v2 presents significant challenges for visual odometry due to its rigorous motion and extreme illumination changes.
The traditional method, ORB-SLAM, fails on most trajectories.
Our approach improves $61.9\%$ in $t_{\mathrm{rel}}$ compared DroidSLAM. Notably, on trajectory H00, which simulates the lunar surface shown in \fref{fig:qualitative}, our model demonstrates a remarkable $82.4\%$ decrease in $t_{\mathrm{rel}}$ and achieves the lowest $r_{\mathrm{rel}}$ among all baseline methods.

\textbf{KITTI Dataset} \;
To further validate our robustness and consistency in outdoor, large-scale trajectory with dynamic objects, we evaluate our system on the KITTI \cite{kitti2012} dataset.
Our method, which relies solely on two-frame pose optimization, shows commendable performance, ranked behind the ORB-SLAM3, a full visual SLAM system, on $t_{\mathrm{rel}}$. 
Our system significantly outperforms other visual odometry approaches in $t_{\mathrm{rel}}$, demonstrating a 53.3\% reduction in relative translation error. 
The degraded performance of $r_{\mathrm{rel}}$ may be attributed to the lack of multi-frame optimization. 

\subsection{Qualitative Analysis and Ablation Study}

\begin{figure*}
    \centering
    \captionsetup{font=small}
    \includegraphics[width=\linewidth]{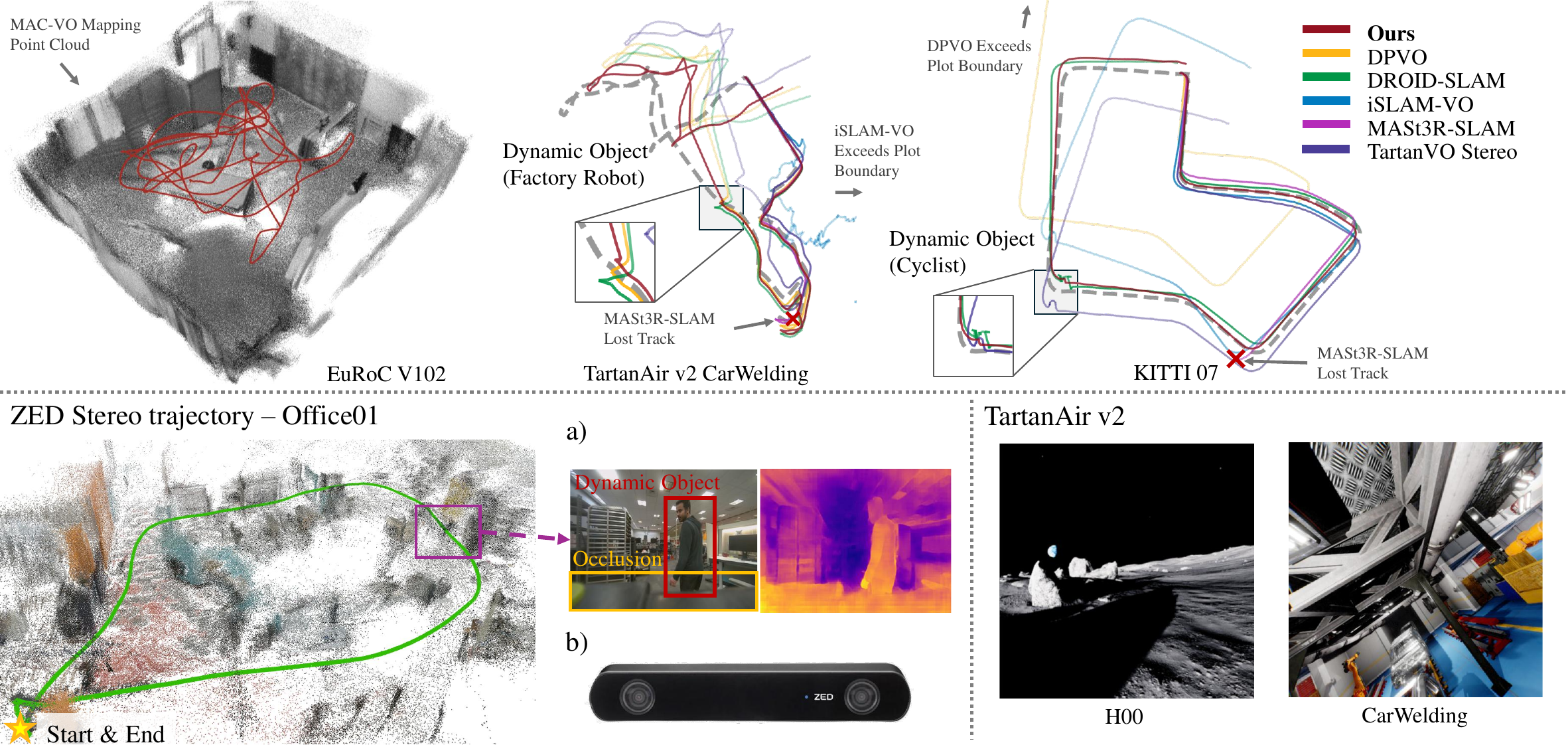}
    \caption{\textbf{Top:} Trajectories estimated by our model and baselines. We highlight the segments where dynamic objects interrupt the VO. \textbf{Bottom left:} We collect data in the office using our own payload with the ZED-X camera. The data contains multiple dynamic objects and visual occlusions.
    % Our method demonstrates robustness against dynamic objects and visual occlusions in the images.
    \textbf{Bottom right:} samples of TartanAir v2 test dataset, which simulates the exotic lunar environment.}
    \label{fig:qualitative}
    \vspace{-5pt}
\end{figure*}

In addition to quantitative evaluations, we conduct qualitative evaluations of our proposed system across multiple datasets including EuRoC, KITTI, and TartanAir v2, supplemented by manually collected data using a ZED-X stereo camera.
MASt3R-SLAM struggles on many trajectories of KITTI and TartanAir, primarily due to incorrect loop closure detections from visual ambiguities.
Our model, even without multi-frame optimization, achieves top-tier performance and exhibits fewer glitches than baseline methods. 
As demonstrated in the \fref{fig:qualitative}, our method produces smoother trajectories and superior pose estimation precision. \fref{fig:qualitative} a) presents that our model correctly identifies the region occluded by the mounted platform and dynamic objects in the scene and assigns a high uncertainty score to these regions.

\begin{table}[H]
\centering
    \captionsetup{font=small}
    \caption{Ablation study. Detailed results in \aref{appendix:AdditionalAblation}.}
    \resizebox{\linewidth}{!}{
    \begin{tabular}{lcccc}
        \toprule
        Dataset & \multicolumn{2}{c}{TartanAir v2 Hard} & \multicolumn{2}{c}{TartanAir v2 Easy}\\
                & $t_{\mathrm{rel}}$ & $r_{\mathrm{rel}}$ & $t_{\mathrm{rel}}$ & $r_{\mathrm{rel}}$\\
        \midrule
        \textbf{Module Ablation}\\
        \hspace{3mm} w/o CovKP \& CovOpt  & .0743 & .5221 & .0521 & .2799 \\
        \hspace{3mm} w/o CovOpt           & .0679 & .3776 & .0511 & .2367 \\
        \hspace{3mm} w/o CovKP            & \underline{.0188} & .2347 & \underline{.0066} & .0808 \\
        \midrule
        \textbf{Covariance Ablation}      & \\
        \hspace{3mm} DiagCov              & .0461 & .3023 & .0277 & .1548 \\
        \hspace{3mm} Scale Agnostic       & .0204 & \underline{.2321} & .0086 & \underline{.0764} \\
        % \hspace{3mm} Intra-frame Agnostic & 0.0208 & 0.1880 & 0.0077 & 0.0750 \\
        % \yuheng{Diag + Scale Agnostic}\\
        \midrule
        \textbf{Ours}                     & \textbf{.0141} & \textbf{.1429} & \textbf{.0051} & \textbf{.0670}\\
        \bottomrule
    \end{tabular}
    }
    \label{tab: AblationStudy}
    \vspace{-5pt}
\end{table}
\begin{figure}[H]
    \centering
    \captionsetup{font=small}
    \includegraphics[width=.8\linewidth]{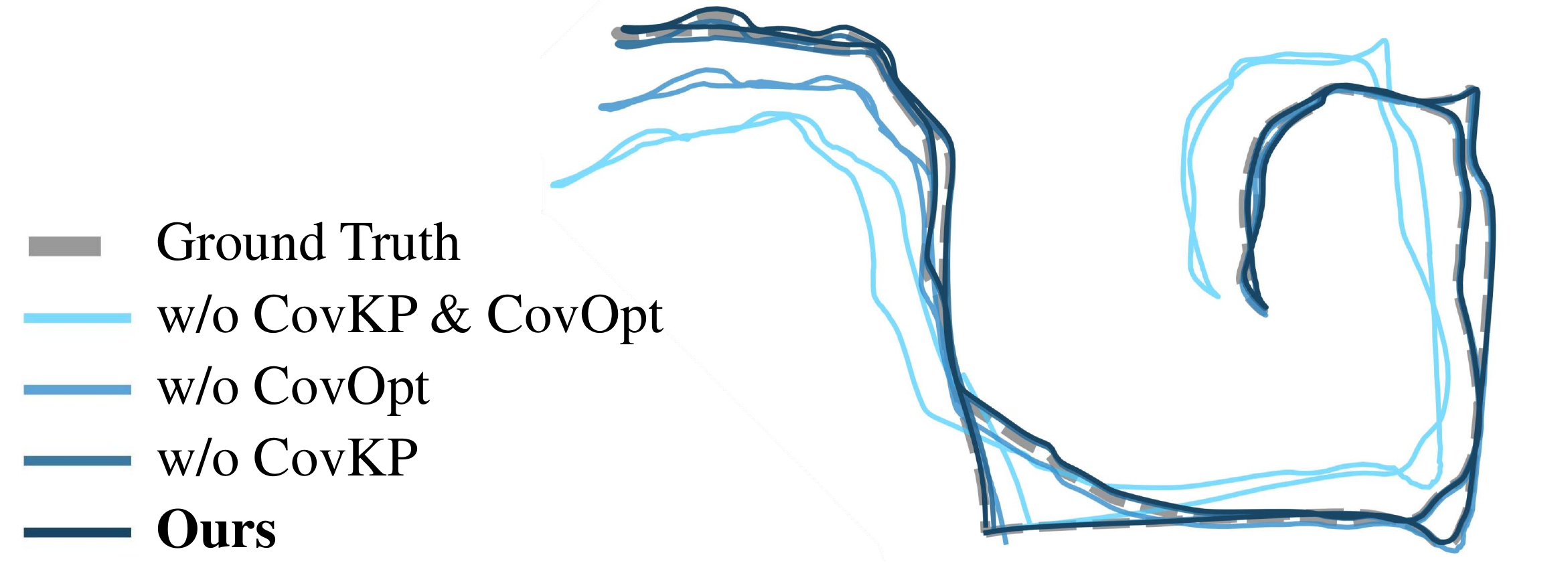}
    \caption{Results of various ablation setups on the H01 of TartanAirv2.}
    \label{fig: Ablation}
    \vspace{-5pt}
\end{figure}

\textbf{Ablation Study}\;\;
\label{sect:AblationStudy}
In \tref{tab: AblationStudy}, we first perform the ablation study on each module of MAC-VO including 
(a. \textit{w/o CovKP \& CovOPT}) with random keypoint selector and identity covariance matrix. 
(b. \textit{w/o CovOpt}) replace the metrics-aware covariance model with the identity covariance model.
(c. \textit{w/o CovKP}) replace the proposed keypoint selector with a random keypoint selector.
To demonstrate the necessity of scale consistency and off-diagonal terms in our covariance model, we run the ablation study with different configurations: 
(\rom{1}. \textit{DiagCov}) remove off-diagonal terms in Eq. \ref{eq: Covariance terms}, 
(\rom{2}. \textit{Scale-agnostic}) normalize the covariance model by the average determinants of covariance matrices of each frame. 
% (\rom{3}. \textit{DiagCov \& Scale Agnostic}) using a normalized diagonal covariance model.

\begin{table}[!t]
\centering
\caption{Time spent on each module of MAC-VO under different optimization with image resolution of $640\times 640$.}
\vspace{-2mm}
\resizebox{\linewidth}{!}{\begin{tabular}{lccccc}
\toprule
   & \textbf{Raw} & \textbf{TRT}$^\star$ & \textbf{MP$^\dagger$ + TRT$^\star$} & \textbf{MAC-VO Fast$^\ddagger$} \\
  \midrule
  \textbf{Modules (ms)}\\
  \hspace{2mm} Frontend Network  & 401.9 & 239.3 & 239.9 & 81.6 \\
  \hspace{2mm} Optimization      & 53.4  & 57.5  & -     & - \\
  \hspace{2mm} Motion Model      & 5.2   & 5.1   & 5.3   & 5.2 \\
  \hspace{2mm} Keypoint Selector & 0.7   & 0.6   & 0.6   & 0.5 \\
  \hspace{2mm} Covariance Model  & 0.6   & 0.7   & 0.7   & 0.7 \\
  \midrule
    \textbf{Overall (fps)} & 2.15 & 3.25 & 3.96 & \textbf{10.57} \\
  \bottomrule
  \multicolumn{5}{l}{\small$^\star$ TRT: TensorRT framework - \href{https://developer.nvidia.com/tensorrt}{https://developer.nvidia.com/tensorrt}
  }\\
  \multicolumn{5}{l}{\small$^\dagger$ MP: multi-processing the PGO in parallel with the matching network.
  }\\
  \multicolumn{5}{l}{\small$^\ddagger$
  MAC-VO Fast: utilizes half-precision number (\texttt{bf16}) and light-weight model.}\\
\end{tabular}}
\label{tab:RuntimeAnalysis}
\vspace{-10pt}
\end{table}

\subsection{Runtime Analysis}

The runtime analysis shown in \tref{tab:RuntimeAnalysis} uses the platform with \texttt{AMD Ryzen 9 5950X} CPU and \texttt{NVIDIA 3090 Ti} GPU.
We also introduce a fast mode (MAC-VO Fast) that utilizes half-precision number in the network inference to enhance efficiency.
This mode also speeds up the memory decoder network by reducing the number of iterative updates from 12 to 4. 
The fast mode performs at 10.5 fps (frames per second) with 70\% of the performance of the original MAC-VO. More details are included in \aref{appendix:RuntimeAnalysis}.

\section{Conclusion \& Discussion}

This paper proposes MAC-VO, a learning-based stereo VO method that outperforms most visual odometry and even SLAM algorithms on challenging datasets.
In our current work, the model focuses on the two-frame pose optimization. 
We believe our accuracy will be further benefit from bundle adjustment, multi-frame optimization, and loop closure. 
Additionally, we plan to apply our metrics-aware covariance model in multi-sensor fusion, such as with IMUs.

% \section{Acknowledgement}

% This work was supported by the DSTA under contract number \#DST000EC124000205.

\clearpage
\bibliographystyle{IEEEtran}
\bibliography{bibliography/papers, bibliography/yuheng}

\clearpage
\appendix
\section{Covariance Model Formulation}
\subsection{Depth Uncertainty From Disparity Uncertainty}
\label{appendix:depthcov}

% In this section, we will present the mathematical formulation for estimating the distribution of depth $\mathbf{d}$ on a single pixel given the estimated distribution of disparity $\Disparity \sim \mathcal{N}(\mu_{\Disparity}, \sigma_\Disparity^2)$.
In this section, we present the formulation for estimating the distribution of depth $\mathbf{d}$ on a single pixel given the estimated distribution of disparity $\Disparity \sim \mathcal{N}(\mu_{\Disparity}, \sigma_\Disparity^2)$.

% Following the pinhole camera model, given the disparity of a pixel point on a pair of rectified cameras with baseline $b$, the depth can be calculated as $\mathbf{d} = bf_x/\Disparity$. 
% Since we assumed $\Disparity$ follows the Gaussian distribution, the probability density of $\Disparity = 0$ is non-zero, which leads to an ill-defined distribution of $\mathbf{d}$.
Following the pinhole camera model, the depth is calculated as $\mathbf{d} = bf_x/\Disparity$ where the camera baseline is $b$. 
% Since 
Since $\Disparity$ can be zero, the distribution of $\mathbf{d}$ may be ill-defined.
% Since we assume $\Disparity$ follows the Gaussian distribution, the probability density of $\Disparity = 0$ is non-zero, which leads to an ill-defined distribution of $\mathbf{d}$.
To fix this, we employ the first-order Taylor expansion to approximate $\mu_d$ and $\sigma_d^2$ such that $\mathbf{d} \sim \mathcal{N}(\mu_d, \sigma_d^2)$.

% Assuming that the variance of disparity is $\sigma_{\Disparity}^2 = (\gamma \mu_\Disparity)^2$ for some sufficiently small $\gamma$ such that the probability for $\Disparity <= 0$ can be ignored. 

We assume the variance of disparity $\sigma_{\Disparity}^2 = (\gamma \mu_\Disparity)^2$ for some sufficiently small $\gamma > 0$ such that the probability of $\Disparity <= 0$ is negligible. Based on this assumption, we have
\begin{equation}
    \begin{aligned}
        \mu_\mathbf{d} &= \mathbb{E}\Biggl[ \frac{bf_x}{\mu_\Disparity} + \left(\frac{d}{d \mu_\Disparity}\frac{bf_x}{\mu_\Disparity}\right)(\Disparity - \mu_\Disparity) \Biggr.\\
        &+ \Biggl.\left(\frac{d^2}{d \mu_\Disparity^2}\frac{bf_x}{\mu_\Disparity}\right)\frac{(\Disparity - \mu_\Disparity)^2}{2} + \cdots \Biggr]\\
            &\approx \mathbb{E}\left[ \frac{bf_x}{\mu_\Disparity} \right] - \frac{1}{\mu_\Disparity^2}\mathbb{E}[(\Disparity - \mu_\Disparity)] \\
            &= \frac{bf_x}{\mu_\Disparity} - 0 =  \frac{bf_x}{\mu_\Disparity} 
    \end{aligned}
\end{equation}
Similarly, $\sigma_{\mathbf{d}}^2$ can be expressed as
\begin{equation}
    \begin{aligned}
        \sigma_\mathbf{d}^2 &= \mathrm{Var}\Biggl[ \frac{bf_x}{\mu_\Disparity} + \left(\frac{d}{d \mu_\Disparity}\frac{bf_x}{\mu_\Disparity}\right)(\Disparity - \mu_\Disparity) \Biggr.\\
        &+ \Biggl.\left(\frac{d^2}{d \mu_\Disparity^2}\frac{bf_x}{\mu_\Disparity}\right)\frac{(\Disparity - \mu_\Disparity)^2}{2} + \cdots \Biggr]\\
            &\approx \mathrm{Var}\left[ \frac{bf_x}{\mu_\Disparity} + \left(\frac{d}{d \mu_\Disparity}\frac{bf_x}{\mu_\Disparity}\right)(\Disparity - \mu_\Disparity) \right] \\
            &= \left(\frac{bf_x}{\mu_\Disparity^2}\right)^2 \cdot \sigma_\Disparity^2 = \frac{(bf_x\gamma)^2}{\mu_\Disparity^2}
    \end{aligned}
\end{equation}

Therefore, the approximation of \textit{depth uncertainty} from disparity uncertainty is expressed as follows:
\begin{equation}
    D \sim \mathcal{N}\left(\frac{bf_x}{\mu_\Disparity},  \frac{(bf_x\gamma)^2}{\mu_\Disparity^2} \right)
\end{equation}

\begin{figure}[htbp]
    \centering
    \includegraphics[width=\linewidth]{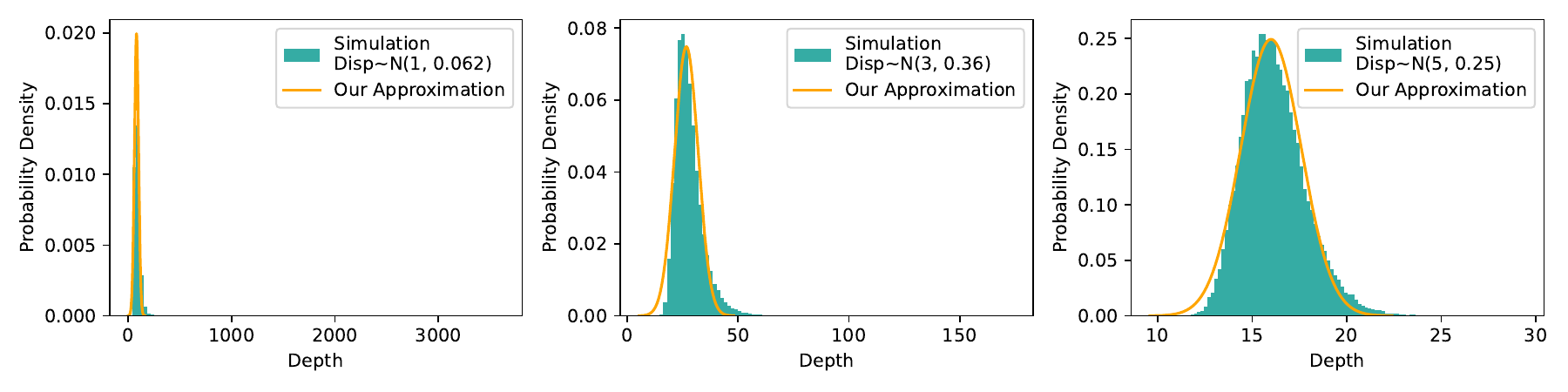}
    \caption{Result of Monte Carlo simulation on depth distribution. Despite the skewness in the simulated distribution, our approximation matches the simulation. As error rate $\gamma$ increases and average disparity $\Disparity$ approaches $0$ (from right to left), the depth distribution becomes more skewed and the quality of approximation decreases.}
    \label{fig:Montecarlo}
\end{figure}

Monte Carlo simulation indicates that for $\gamma < 0.3$, the error of the aforementioned approximation is acceptable, as shown in \fref{fig:Montecarlo}.

\subsection{Uncertainty Correction After Keypoint Matching}
\label{appendix:uncertainty_correction}

Let $q_{i, t}$ be the $i$-th keypoint on the camera plane at time $t$. Given the estimated optical flow $\mathbf{f}$ at $q_{i, t}$, the matched keypoint at time $t+1$ is defined as $q_{i, t+1} = q_{i, t} + \mathbf{f}_{i, t}$. Since $\mathbf{f}_{i, t}$ follows the gaussian distribution $\mathcal{N}(\hat{f}_{i, t}, \hat{\Sigma}_{i, t})$, $q_{i, t+1}$ is a random variable following distribution of $\mathcal{N}(q_{i, t} + \hat{f}_{i, t}, \hat{\Sigma}_{i, t})$.

Let $\varphi_{i, t}$ be a 2D Gaussian filter with covariance matrix $\hat{\Sigma}_{i, t}$, the probability for matched keypoint on some pixel $j$ is then expressed as $(\varphi_{i, t})_j$. Let $d_j$ denote the estimated depth at pixel $j$, then the average depth for pixel $q_{i, t+1}$ weighted by $\varphi$ is expressed as
\begin{equation}
    \mu_{d_{i, t}} = \sum_{j}{(\varphi_{i, t})_j \cdot d_j},
\end{equation}
and the estimated variance of depth of $q_{i, t+1}$ is calculated as weighted variance
\begin{equation}
    \sigma_{d_{i, t}}^2 = \sum_j{(\varphi_{i, t})_j \cdot (d_j - \mu_{d_{i, t}})^2}.
\end{equation}
% It is also possible to model the depth of the matched point as the mixture of Gaussian distribution. However, experiment results indicate that using a mixture of Gaussian distribution only provides a minimal performance improvement. We therefore adopt the straightforward weighted variance approach to estimate depth uncertainty.
We could also model the depth of the matched point using a mixture of Gaussian distributions, but experiments show that this offers only a minimal performance improvement. Therefore, we use the straightforward weighted variance method to estimate depth uncertainty.

\begin{figure}
    \centering
    \includegraphics[width=0.7\linewidth]{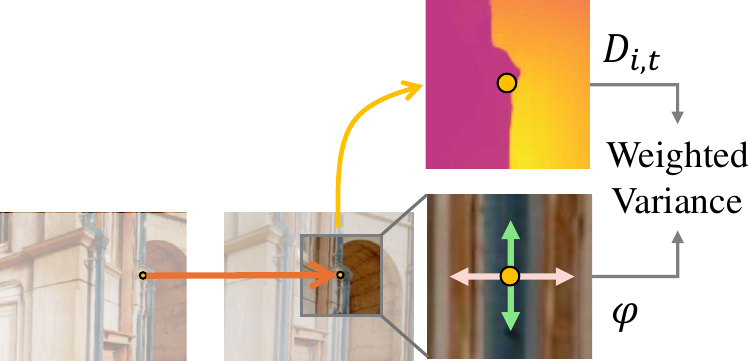}
    \caption{Estimate depth variance of the matched point under the presence of matching uncertainty}
    \label{fig:depth-patch-estimate-cov}
\end{figure}

\subsection{Projecting 2D Uncertainty to Spatial Covariance}
\label{appendix:Project2DTo3D}

% Let $\mathbf{u}_{i, t} \sim \mathcal{N}(u_{i, t}, \sigma_{u_{i, t}}^2)$, $\mathbf{v}_{i, t} \sim \mathcal{N}(v_{i, t}, \sigma_{v_{i, t}}^2)$, and $\mathbf{d}_{i, t} \sim \mathcal{N}(d_{i, t}, \sigma_{d_{i, t}}^2)$, we will derive the distribution of 3D point under camera coordinate ${}^c\mathbf{p}_{i, t} = [\mathbf{x}_{i, t}, \mathbf{y}_{i, t}, \mathbf{z}_{i, t}]^\top \sim \mathcal{N}({}^cp_{i, t}, {}^c\Sigma_{i, t})$.
Let $\mathbf{u}_{i, t} \sim \mathcal{N}(u_{i, t}, \sigma_{u_{i, t}}^2)$, $\mathbf{v}_{i, t} \sim \mathcal{N}(v_{i, t}, \sigma_{v_{i, t}}^2)$, and $\mathbf{d}_{i, t} \sim \mathcal{N}(d_{i, t}, \sigma_{d_{i, t}}^2)$, we derive the distribution of 3D point under camera coordinate ${}^c\mathbf{p}_{i, t} = [\mathbf{x}_{i, t}, \mathbf{y}_{i, t}, \mathbf{z}_{i, t}]^\top \sim \mathcal{N}({}^cp_{i, t}, {}^c\Sigma_{i, t})$.

Recall that the relationship between pixel coordinate $(\mathbf{u}_{i, t}, \mathbf{v}_{i, t})$, depth $\mathbf{d}_{i, t}$ and 3D coordinate $\mathbf{x}_{i, t}, \mathbf{y}_{i, t}, \mathbf{z}_{i, t}$ is depicted as
\begin{equation}
    \mathbf{x}_{i, t} = \frac{(\mathbf{u}_{i, t} - c_x)\mathbf{d}_{i, t}}{f_x},\quad \mathbf{y}_{i, t} = \frac{(\mathbf{v}_{i, t} - c_y)\mathbf{d}_{i, t}}{f_y},\quad \mathbf{z}_{i, t} = \mathbf{d}_{i, t}
\end{equation}

Assume $\mathbf{u}_{i, t}$, $\mathbf{v}_{i, t}$, $\mathbf{d}_{i, t}$ are independent to each other, we have $\mathrm{Var}(\mathbf{u}_{i, t}\mathbf{d}_{i, t}) = (\sigma_{u_{i, t}}^2 + d_{i, t}^2)(\sigma_{d_{i, t}}^2 + u_{i, t}^2) - u_{i, t}^2d_{i, t}^2$  \cite{ExactVarianceProduct}. Based on this expression of variance, it follows that
\begin{equation}
\begin{aligned}
    % \sigma_{x_{i, t}}^2 &= \mathrm{Var}\left( \frac{\mathbf{u}_{i, t}\mathbf{d}_{i, t}}{f_x} - \frac{c_x \mathbf{d}_{i, t}}{f_x} \right) = \frac{\mathrm{Var}(\mathbf{u}_{i, t}\mathbf{d}_{i, t})}{f_x^2} + \frac{c_x^2 \mathrm{Var}(\mathbf{d}_{i, t})}{f_x^2} \\
    %     &= \frac{(\sigma_{u_{i, t}}^2 + d_{i, t}^2)(\sigma_{d_{i, t}}^2 + u_{i, t}^2) - u_{i, t}^2d_{i, t}^2 + c_x^2\sigma_{d_{i, t}}^2}{f_x^2}\\
    \sigma_{x_{i, t}}^2 &= \mathrm{Var}\left(\frac{(\mathbf{u}_{i, t} - c_x)\mathbf{d}_{i, t}}{f_x} \right) = \frac{\mathrm{Var}((\mathbf{u}_{i, t} - c_x)\mathbf{d}_{i, t})}{f_x^2}\\
        &= \frac{\sigma_{u_{i, t}}^2\sigma_{d_{i, t}}^2 + \sigma_{u_{i, t}}^2\mathbf{d}_{i, t}^2 + (\mathbf{u}_{i, t} - c_x)^2\sigma_{d_{i, t}}^2}{f_x^2}\\
    % \sigma_{y_{i, t}}^2 &= \mathrm{Var}\left( \frac{\mathbf{v}_{i, t}\mathbf{d}_{i, t}}{f_y} - \frac{c_y \mathbf{d}_{i, t}}{f_y} \right) = \frac{\mathrm{Var}(\mathbf{v}_{i, t}\mathbf{d}_{i, t})}{f_y^2} + \frac{c_y^2 \mathrm{Var}(\mathbf{d}_{i, t})}{f_y^2} \\
    %     &= \frac{(\sigma_{v_{i, t}}^2 + d_{i, t}^2)(\sigma_{d_{i, t}}^2 + v_{i, t}^2) - v_{i, t}^2d_{i, t}^2 + c_y^2\sigma_{d_{i, t}}^2}{f_y^2}\\
    \sigma_{y_{i, t}}^2 &= \mathrm{Var}\left(\frac{(\mathbf{v}_{i, t} - c_y)\mathbf{d}_{i, t}}{f_y} \right) = \frac{\mathrm{Var}((\mathbf{v}_{i, t} - c_y)\mathbf{d}_{i, t})}{f_y^2}\\
        &= \frac{\sigma_{v_{i, t}}^2\sigma_{d_{i, t}}^2 + \sigma_{v_{i, t}}^2\mathbf{d}_{i, t}^2 + (\mathbf{v}_{i, t} - c_y)^2\sigma_{d_{i, t}}^2}{f_y^2}\\
    \sigma_{z_{i, t}}^2 &= \sigma_{d_{i, t}}^2
\end{aligned}
\end{equation}

% Since the covariance between two random variables is defined as $\mathrm{Cov}(X, Y) = \mathbb{E}[XY] - \mathbb{E}[X]\mathbb{E}[Y]$, 
Under the assumption that $\mathbf{u}_{i, t}, \mathbf{v}_{i, t}, \mathbf{d}_{i, t}$ are independent to each other, we derive the covariance between $\mathbf{x}_{i, t}$, $\mathbf{y}_{i, t}$ and $\mathbf{z}_{i, t}$ as:
\begin{equation}
    \begin{aligned}
        \mathrm{Cov}(\mathbf{x}_{i, t}, \mathbf{y}_{i, t}) &= \mathrm{Cov}\left(\frac{(\mathbf{u}_{i, t} - c_x)\mathbf{d}_{i, t}}{f_x}, \frac{(\mathbf{v}_{i, t} - c_y)\mathbf{d}_{i, t}}{f_y}\right)\\
            &= \mathbb{E}\left[\frac{\mathbf{d}_{i, t}^2 (\mathbf{u}_{i, t} - c_x) (\mathbf{v}_{i, t} - c_y)}{f_xf_y}\right] \\
            &- \mathbb{E}\left[\frac{(\mathbf{u}_{i, t} - c_x)\mathbf{d}_{i, t}}{f_x}\right]\mathbb{E}\left[\frac{(\mathbf{v}_{i, t} - c_y)\mathbf{d}_{i, t}}{f_y}\right]\\
            &= \frac{(\mathbb{E}[\mathbf{d}_{i, t}^2] - \mathbb{E}[\mathbf{d}_{i, t}]^2)\mathbb{E}[\mathbf{u}_{i, t} - c_x]\mathbb{E}[\mathbf{v}_{i, t} - c_y]}{f_xf_y}\\
            &= \frac{\sigma_{d_{i, t}}^2}{f_xf_y}(u_{i, t} - c_x) (v_{i, t} - c_y)
    \end{aligned}
\end{equation}
and
\begin{equation}
    \begin{aligned}
    \mathrm{Cov}(\mathbf{x}_{i, t}, \mathbf{z}_{i, t}) &= \mathrm{Cov}\left( \frac{(\mathbf{u}_{i, t} - c_x)\mathbf{d}_{i, t}}{f_x}, \mathbf{d}_{i, t} \right) \\
    &= \frac{(\mathbb{E}[\mathbf{d}_{i, t}^2] - \mathbb{E}[\mathbf{d}_{i, t}]^2)\mathbb{E}[\mathbf{u}_{i, t}]}{f_x} = \frac{\sigma_{d_{i, t}}^2}{f_x}(u_{i, t} - c_x)\\
    \mathrm{Cov}(\mathbf{y}_{i, t}, \mathbf{z}_{i, t}) &= \mathrm{Cov}\left( \frac{(\mathbf{v}_{i, t} - c_y)\mathbf{d}_{i, t}}{f_y}, \mathbf{d}_{i,t} \right)\\
    &= \frac{(\mathbb{E}[\mathbf{d}_{i, t}^2] - \mathbb{E}[\mathbf{d}_{i, t}]^2)\mathbb{E}[\mathbf{v}_{i, t}]}{f_y} = \frac{\sigma_{d_{i, t}}^2}{f_y}(v_{i, t} - c_y)
    \end{aligned}
\end{equation}

\begin{figure}[h!]
    \centering
    \includegraphics[width=\linewidth]{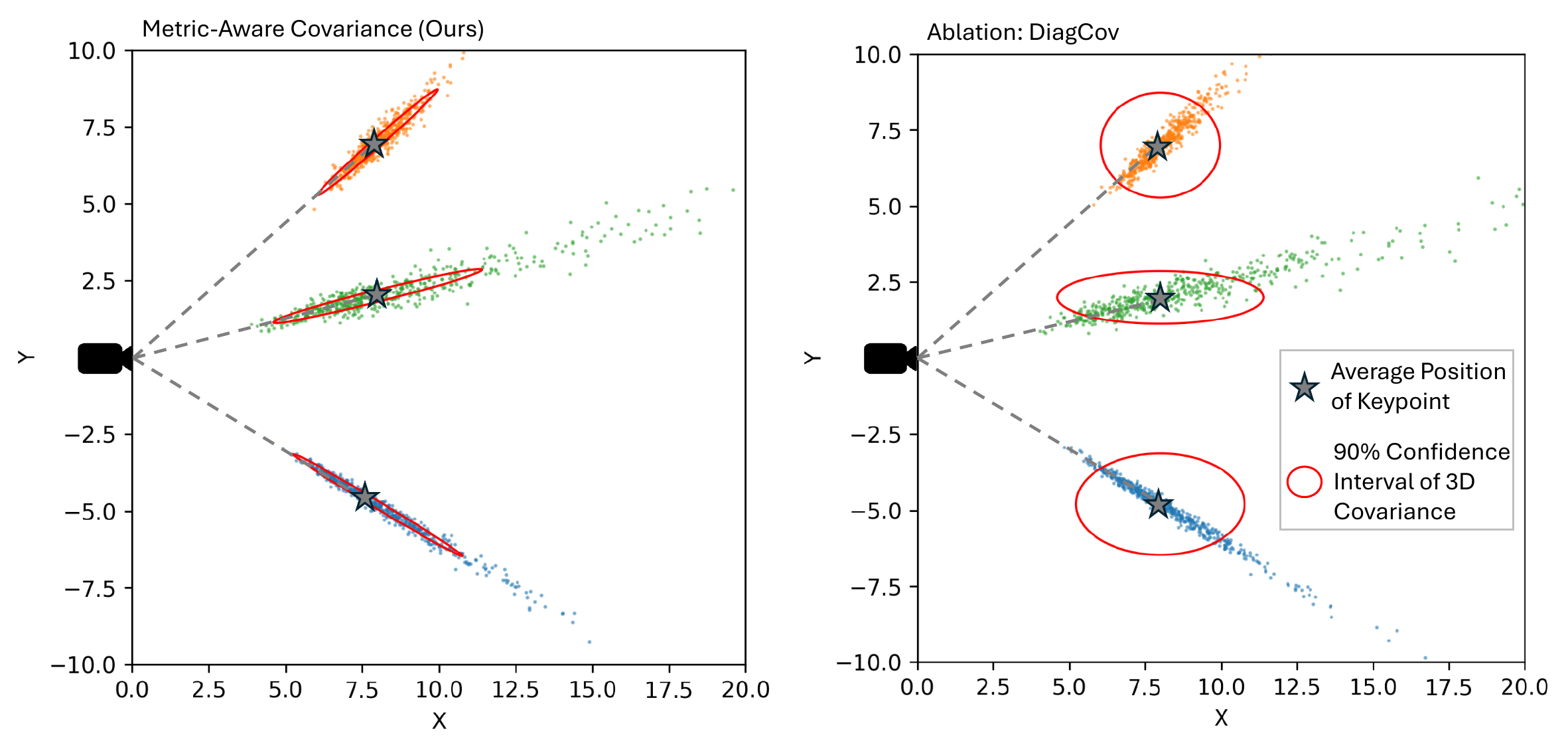}
    \caption{Comparison of proposed 3D covariance (\textbf{left}) and the diagonal covariance matrix (\textbf{right}, DiagCov in ablation study). Our method captures the uncertainty of keypoints significantly better than the diagonal covariance matrix.}
    \label{fig: DiagonalCovarianceVisualization}
\end{figure}

\fref{fig: DiagonalCovarianceVisualization} visualize the distribution of keypoints in 3D space via Monte Carlo and the 90\% confidence interval of estimated distribution, confirming the necessity of off-diagonal terms.

\section{Additional Results}

\subsection{Remaining Results on EuRoC, TartanAirv2, and KITTI}
\label{appendix:AdditionalResults}

See \tref{tab:EuRoCEven}, \tref{tab: TartanAirEasy}, and \tref{tab: KITTI Odd} for data.

\begin{table*}[ht]
\centering
\captionsetup{font=small}
\caption{Performance comparison of different methods on the EuRoC Dataset. Only even-ordered trajectory is shown here, see \tref{tab:EuRoCOdd} for the remaining results.}
\resizebox{.8\linewidth}{!}{
\begin{tabular}{lcccccccccccc}
\toprule
 Trajectory & \multicolumn{2}{c}{{MH02}} & \multicolumn{2}{c}{{MH04}} & \multicolumn{2}{c}{{V101}} & \multicolumn{2}{c}{V103} & \multicolumn{2}{c}{V202}\\
\cmidrule{2-11}
   &  $t_{\mathrm{rel}}$ & $r_{\mathrm{rel}}$ & $t_{\mathrm{rel}}$ & $r_{\mathrm{rel}}$ & $t_{\mathrm{rel}}$ & $r_{\mathrm{rel}}$ & $t_{\mathrm{rel}}$ & $r_{\mathrm{rel}}$ & $t_{\mathrm{rel}}$ & $r_{\mathrm{rel}}$\\
\midrule
\textbf{SLAM} \\
\hspace{3mm} \stereoicon~ORB-SLAM 3                & 0.0036 & 0.0495   &  0.0061  & 0.0501   &  0.0049  & 0.0888   &  0.0137  & 0.2669   &  0.0090  & 0.1528\\
\hspace{3mm} \stereoicon~DROID-SLAM$^\star$         & \textbf{0.0012} & \textbf{0.0169} & 0.0031   & \textbf{0.0224} & \textbf{0.0024} & \underline{0.0314} & 0.0036 & \underline{0.0642} & \textbf{0.0017} & \textbf{0.0399}\\
\hspace{3mm} \monoicon~MASt3R-SLAM$^{\star}$ & 0.0138 & 0.5493 & 0.0274 & 0.8890 & 0.0130 & 0.5808 & 0.0383 & 1.3178 & 0.0334 & 0.6818 \\
\midrule
\textbf{VO} \\
\hspace{3mm} \monoicon~TartanVO$^{\star}$  & 0.0172 & 0.0621 & 0.0213 & 0.0681 & 0.0124 & 0.0756 & 0.0263 & 0.1552 & 0.0171 & 0.1251\\
\hspace{3mm} \stereoicon~TartanVO                  & 0.0289 & 0.5037 & 0.0501 & 0.5400 & 0.0224 & 0.5322 & 0.0351 & 1.3127 & 0.0361 & 1.1607\\
\hspace{3mm} \stereoicon~iSLAM-VO                   & 0.0041 & 0.0620 & 0.0082 & 0.0682 & 0.0041 & 0.0756 & 0.0088 & 0.1554 & 0.0078 & 0.1252\\
\hspace{3mm} \monoicon~DPVO$^{\star}$      & 0.0014 & 0.0212 & \underline{0.0029} & \underline{0.0264} & \underline{0.0026} & 0.0405   & \underline{0.0033} & 0.0662   & 0.0022 & 0.0493\\
% \midrule
\rowcolor{ourgray}
\hspace{3mm} \stereoicon~\textbf{Ours}                           & \underline{0.0013} & \underline{0.0199} & \textbf{0.0028} & 0.0273   & \textbf{0.0024} & \textbf{0.0304} & \textbf{0.0032} & \textbf{0.058} & \underline{0.0018} & \underline{0.0406}\\
\bottomrule
\multicolumn{10}{l}{
\monoicon \hspace{.5mm} Monocular method.
\quad
\stereoicon \hspace{.5mm} Stereo method.
\quad
$^\star$ Scale-aligned with ground truth.} \\
\end{tabular}
}
\label{tab:EuRoCEven}
\end{table*}

\begin{table*}[htbp]
\centering
\captionsetup{font=small}
\caption{Performance comparison on the TartanAir v2 Easy Dataset.}
% \vspace{-2mm}
\resizebox{\linewidth}{!}{
\begin{tabular}{lcccccccccccccccc}
\toprule
 Trajectory & \multicolumn{2}{c}{{E00}} & \multicolumn{2}{c}{{E01}}  & \multicolumn{2}{c}{{E02}} & \multicolumn{2}{c}{{E03}}  & \multicolumn{2}{c}{{E04}} & \multicolumn{2}{c}{{E05}}  & \multicolumn{2}{c}{{E06}} & \multicolumn{2}{c}{Avg.} \\
\cmidrule{2-17}
   &  $t_{\mathrm{rel}}$ & $r_{\mathrm{rel}}$ & $t_{\mathrm{rel}}$ & $r_{\mathrm{rel}}$ & $t_{\mathrm{rel}}$ & $r_{\mathrm{rel}}$ & $t_{\mathrm{rel}}$ & $r_{\mathrm{rel}}$ & $t_{\mathrm{rel}}$ & $r_{\mathrm{rel}}$ & $t_{\mathrm{rel}}$ & $r_{\mathrm{rel}}$ & $t_{\mathrm{rel}}$ & $r_{\mathrm{rel}}$ & $t_{\mathrm{rel}}$ & $r_{\mathrm{rel}}$ \\
\midrule
\textbf{SLAM} \\
\hspace{3mm} \stereoicon~ORB-SLAM3                  & - & - & .1019 & 2.349 & - & - & - & - & - & - & - &- & - & - & .1019 & 2.349 \\
\hspace{3mm} \monoicon~MASt3R-SLAM$^{\star}$ & .3268 & 2.892 & .3124 & 2.921 & - & - & - & - & .1872 & 2.948 & - & - & .2018 & 3.033 & .2570 & .2948 \\
\hspace{3mm} \stereoicon~DROID-SLAM$^\star$          & \underline{.0077} & \textbf{.0144} & \textbf{.0025} & \textbf{.0199} & \underline{.0063} & \textbf{.0409} & \textbf{.0049} & \textbf{.0251} & \textbf{.0009} & \textbf{.0147} & \underline{.0031} & \underline{.0463} & \textbf{.0016} & \textbf{.0235} & \textbf{.0039} & \textbf{.0264} \\
\midrule
\textbf{VO} \\
\hspace{3mm} \monoicon~TartanVO$^{\star}$    & .0532 & .3237 & .0937 & .4750 & .1066 & .6048 & .0756 & .2230 & .1114 & .4032 & .0862 & .3185 & .1373 & .6620 & .0949 & .4300 \\
\hspace{3mm} \stereoicon~TartanVO                    & .0505 & .1334 & .0322 & .2078 & .0237 & .1105 & .0303 & .1417 & .0173 & .1681 & .0279 & .3828 & .0218 & .2329 & .0291 & .1967 \\
\hspace{3mm} \stereoicon~iSLAM-VO                    & .0656 & .2873 & .0456 & .3853 & .0359 & .2234 & .0508 & .2635 & .0268 & .3201 & .0464 & .6624 & .0362 & .4606 & .0439 & .3718 \\
\hspace{3mm} \monoicon~DPVO$^{\star}$       & .0113 & \underline{.0187} & \underline{.0047} & \underline{.0249} & .0099 & \underline{.0475} & .0603 & .2064 & \underline{.0044} & \underline{.0177} & .0511 & .0665 & .0189 & .1543 & .0229 & .0766 \\
\rowcolor{ourgray}
\hspace{3mm} \stereoicon~\textbf{Ours}                            & \textbf{.0026} & .0351 & .0124 & .1183 & \textbf{.0031} & .0684 & \underline{.0054} & \underline{.0383} & .0050 & .0413 & \textbf{.0018} & \textbf{.0247} & \underline{.0054} & \underline{.1427} & \underline{.0051} & \underline{.0670} \\
\bottomrule
\multicolumn{17}{l}{
\monoicon \hspace{.5mm} Monocular method.
\quad
\stereoicon \hspace{.5mm} Stereo method.
\quad
$^\star$ Scale-aligned with ground truth.} \\
\end{tabular}
}
\vspace{-5pt}
\label{tab: TartanAirEasy}
\end{table*}

\begin{table*}[ht]
\centering
\captionsetup{font=small}
\caption{Performance comparison of different methods on the KITTI Dataset. Only odd-numbered trajectory is shown here, see \tref{tab:KITTIEven} for the remaining results.}
\resizebox{.8\linewidth}{!}{
\begin{tabular}{lcccccccccc}
\toprule
 Trajectory & \multicolumn{2}{c}{{01}} & \multicolumn{2}{c}{{03}} & \multicolumn{2}{c}{{05}} & \multicolumn{2}{c}{07} & \multicolumn{2}{c}{09} \\
% \cmidrule{2-15}
   &  $t_{\mathrm{rel}}$ & $r_{\mathrm{rel}}$ & $t_{\mathrm{rel}}$ & $r_{\mathrm{rel}}$ & $t_{\mathrm{rel}}$ & $r_{\mathrm{rel}}$ & $t_{\mathrm{rel}}$ & $r_{\mathrm{rel}}$ & $t_{\mathrm{rel}}$ & $r_{\mathrm{rel}}$ \\
\midrule
\textbf{SLAM} \\
\hspace{3mm} \stereoicon~ORB-SLAM 3                 &  \textbf{0.0416}  & \underline{0.0355} &  0.027  & 0.0425   &  0.0161  & 0.0416   &  \textbf{0.0155}  & \underline{0.0385} &  \underline{0.0208}  & 0.0444   \\
\hspace{3mm} \stereoicon~DROID-SLAM$^\star$         &  0.7112  & 0.0406   &  \textbf{0.0182}  & \underline{0.0385} &  \underline{0.0153}  & \underline{0.0353} & 0.0746   & 0.0734   &  0.0214  & \underline{0.0378} \\
\hspace{3mm} \monoicon~MASt3R-SLAM$^{\star}$ & 1.9316 & 0.3442 & 0.5313 & 0.2917 & - & - & 0.3735 & 0.3913 & 0.8414 & 0.5479 \\
\midrule
\textbf{VO} \\
\hspace{3mm} \monoicon~TartanVO$^{\star}$  & 0.6834 & 0.0895 & 0.1234 & 0.0682 & 0.1821 & 0.0761 & 0.2005 & 0.0847 & 0.1704 & 0.1069 \\
\hspace{3mm} \stereoicon~TartanVO                   & 1.1408 & 0.2455 & 0.0477 & 0.0953 & 0.0637 & 0.0821 & 0.0700 & 0.0931 & 0.0990 & 0.1077 \\
\hspace{3mm} \stereoicon~iSLAM-VO                    & 0.2978 & 0.0896 & 0.0507 & 0.0681 & 0.0504 & 0.0758 & 0.0593 & 0.0842 & 0.0660 & 0.1064 \\
\hspace{3mm} \monoicon~DPVO$^{\star}$      & \underline{0.0942} & \textbf{0.0247} & 0.0302 & \textbf{0.0330} & 0.2221 & \textbf{0.0319} & 0.1064 & \textbf{0.0311} & 0.1723 & \textbf{0.0336}\\
% \midrule
\rowcolor{ourgray}
\hspace{3mm} \stereoicon~\textbf{Ours}                           & 0.2663 & 0.1670   & \underline{0.0187} & 0.0504   & \textbf{0.0142} & 0.0466   & \underline{0.0194} & 0.0507 & \textbf{0.0182} & 0.0567 \\
\bottomrule
\multicolumn{10}{l}{
\monoicon \hspace{.5mm} Monocular method.
\quad
\stereoicon \hspace{.5mm} Stereo method.
\quad
$^\star$ Scale-aligned with ground truth.} \\
\end{tabular}
}
\vspace{-10pt}
\label{tab: KITTI Odd}
\end{table*}

\subsection{Robustness Analysis}

See \tref{tab: robustness} for data.

\begin{table}
    \centering
    \captionsetup{font=small}
    \caption{Robustness of systems on TartanAir v2 Hard test dataset demonstrate by the average variance of relative translation and rotation error.}
    \resizebox{\linewidth}{!}{%
    \begin{tabular}{lcccccc}
        \toprule
        Model & DROID-SLAM & iSLAM-VO & TartanVO$^{\star\dagger}$ & TartanVO & DPVO$^{\star\dagger}$ & \textbf{Ours}\\
        \midrule
        Avg. $\sigma^2_{t_{\mathrm{rel}}}$ & \underline{0.072} & 0.383 & 0.169 & 0.107 & 0.318 & \textbf{0.045}\\
        \vspace{-2mm}\\
        Avg. $\sigma^2_{r_{\mathrm{rel}}}$ & \textbf{0.418} & 2.925 & 2.748 & 0.991 & 2.099 & \underline{0.475}\\
        \bottomrule
        \multicolumn{6}{l}{\small{$^\star$ The estimated sequence is scale-aligned with ground truth.}}\\
        \multicolumn{6}{l}{\small{$^\dagger$ Monocular method.}}
    \end{tabular}}
    \label{tab: robustness}
\end{table}

\subsection{Additional Ablation Study}

% We further demonstrate the necessity of metric-aware covariance in our system by performing additional ablation studies on the scale of covariance. Two setups are used to normalize the covariance and remove scale information. 1) The average norm of all observation covariance in a frame is normalized to $1$ (\textit{FrameNorm}); 2) The norm of each observation covariance is normalized to unit vector separately (\textit{PointNorm}).

\label{appendix:AdditionalAblation}

See \tref{tab: Ablation Hard} and \tref{tab: Ablation Easy} for data.

\begin{table}[ht]
\centering
\captionsetup{font=small}
\caption{Performance comparison of different ablation setups on the TartanAir v2 Hard Dataset.}
\resizebox{.9\linewidth}{!}{\begin{tabular}{lccccccc}
\multicolumn{6}{l}{\textbf{Relative Translation Error ($t_{\mathrm{rel}}, \mathrm{m} / \mathrm{frame}$)}}\\
\toprule
Trajectory & H00 & H01 & H02 & H03 & H04 & H05 & H06 \\
\midrule
\textbf{System Modules}\\
\hspace{2mm}w/o CovKP \& CovOpt & 0.136 & 0.037 & 0.058 & 0.098 & 0.027 & 0.071 & 0.092 \\
\hspace{2mm}w/o CovOpt          & 0.171 & 0.033 & 0.061 & 0.106 & 0.023 & 0.034 & 0.048 \\
\hspace{2mm}w/o CovKP           & \underline{0.025} & \underline{0.005} & \underline{0.015} & \underline{0.025} & \textbf{0.005} & 0.029 & 0.028 \\
\midrule
\textbf{Covariance Model}\\
\hspace{2mm}DiagCov         & 0.128 & 0.016 & 0.038 & 0.066 & 0.014 & 0.023 & 0.038 \\
\hspace{2mm}Scale Agnostic  & 0.041 & \textbf{0.004} & 0.018 & 0.027 & \textbf{0.005} & \underline{0.021} & \underline{0.027} \\
% \hspace{2mm}PointNorm     & 0.013 & 0.039 & \underline{0.008} & \underline{0.015} & 0.056 & \underline{0.005} & 0.033 & \underline{0.008} & 0.019 & 0.027 & \underline{0.013} & \underline{0.007} & 0.022 & \underline{0.027} \\
\midrule
\textbf{Ours} & \textbf{0.008} & 0.034 & \textbf{0.005} & \textbf{0.015} & 0.009 & \textbf{0.005} & \textbf{0.020}  \\
\bottomrule
\end{tabular}}

\vspace{3mm}

\resizebox{.9\linewidth}{!}{\begin{tabular}{lccccccc}
\multicolumn{6}{l}{\textbf{Relative Rotation Error ($r_{\mathrm{rel}}, {}^\circ / \mathrm{frame}$)}}\\
\toprule
Trajectory & H00 & H01 & H02 & H03 & H04 & H05 & H06 \\
\midrule
\textbf{System Modules}\\
\hspace{2mm}w/o CovKP \& CovOpt & 0.380 & 0.277 & 0.206 & 0.456 & 0.248 & 0.926 & 1.160 \\
\hspace{2mm}w/o CovOpt          & 0.350 & 0.244 & 0.248 & 0.507 & 0.176 & 0.449 & 0.669 \\

\hspace{2mm}w/o CovKP           & \textbf{0.099} & \textbf{0.042} & \underline{0.076} & \underline{0.247} & \textbf{0.059} & 0.629 & \underline{0.491} \\
\midrule
\textbf{Covariance Model}\\
\hspace{2mm}DiagCov         & 0.333 & 0.126 & 0.192 & 0.404 & 0.137 & \underline{0.379} & 0.544 \\
\hspace{2mm}Scale Agnostic  & 0.173 & \textbf{0.042} & 0.098 & 0.253 & \underline{0.060} & 0.500& 0.498 \\
% \hspace{2mm}PointNorm   & 0.136 & 0.174 & 0.067 & 0.132 & 0.225 & \underline{0.045} & 0.110  & 0.088 & 0.102 & 0.257 & \underline{0.190} & 0.087 & 0.529 & \underline{0.491} \\
\midrule
\textbf{Ours}  & \underline{0.102} & 0.145 & \textbf{0.063} & \textbf{0.078} & 0.141 & \textbf{0.055} & \textbf{0.465} \\
\bottomrule
\end{tabular}}
\label{tab: Ablation Hard}
\end{table}

\begin{table}[ht]
\centering
\captionsetup{font=small}
\caption{Performance comparison of different ablation setups on the TartanAir v2 Easy Dataset.}
\resizebox{.9\linewidth}{!}{\begin{tabular}{lccccccc}
\multicolumn{6}{l}{\textbf{Relative Translation Error ($t_{\mathrm{rel}}, \mathrm{m} / \mathrm{frame}$)}}\\
\toprule
Trajectory & E00 & E01 & E02 & E03 & E04 & E05 & E06\\
\midrule
\textbf{System Modules}\\
\hspace{2mm}w/o CovKP \& CovOpt & 0.116           & 0.049          & 0.045          & 0.058          & 0.018          & 0.035          & 0.044          \\

\hspace{2mm}w/o CovOpt & 0.124          & 0.052          & 0.048          & 0.070          & 0.016          & 0.020           & 0.027          \\
\hspace{2mm}w/o CovKP & \underline{0.010} & \textbf{0.003} & \underline{0.006} & \underline{0.008} & \textbf{0.002} & 0.009          & 0.009          \\
\midrule
\textbf{Covariance Model}\\
\hspace{2mm}DiagCov & 0.069          & 0.025          & 0.025          & 0.037          & 0.009          & 0.012          & 0.017          \\
% \hspace{2mm}PointNorm & \underline{0.004} & 0.023          & \underline{0.004} & 0.010           & 0.013          & \underline{0.004} & 0.005          & \underline{0.005} & 0.007          & 0.009          & 0.007          & \underline{0.003} & 0.006          & \underline{0.008} \\
\hspace{2mm}Scale Agnostic & 0.026          & \underline{0.004} & \underline{0.006} & 0.009          & \textbf{0.002} & \underline{0.005} & \underline{0.008} \\
\midrule
\textbf{Ours} & \textbf{0.003} & 0.012          & \textbf{0.003} & \textbf{0.005} & 0.005          & \textbf{0.002} & \textbf{0.005} \\
\bottomrule
\end{tabular}}

\vspace{3mm}

\resizebox{.9\linewidth}{!}{\begin{tabular}{lccccccc}
\multicolumn{6}{l}{\textbf{Relative Rotation Error ($r_{\mathrm{rel}}, {}^\circ / \mathrm{frame}$)}}\\
\toprule
Trajectory & E00 & E01 & E02 & E03 & E04 & E05 & E06 \\
\midrule
\textbf{System Modules}\\
\hspace{2mm}w/o CovKP \& CovOpt & 0.238          & 0.249          & 0.151          & 0.204          & 0.152          & 0.492          & 0.473          \\
\hspace{2mm}w/o CovOpt & 0.219          & 0.254         & 0.187          & 0.275          & 0.132          & 0.278          & 0.313          \\
\hspace{2mm}w/o CovKP & \underline{0.048} & \textbf{0.031} & \textbf{0.039} & \underline{0.058} & \textbf{0.025} & 0.189          & 0.174          \\
\midrule
\textbf{Covariance Model}\\
\hspace{2mm}DiagCov & 0.189          & 0.127          & 0.106          & 0.167          & 0.085          & 0.179          & 0.210           \\
% \hspace{2mm}PointNorm & 0.050           & 0.069          & 0.042          & 0.105          & 0.138          & 0.034          & \underline{0.027} & 0.089          & 0.052          & 0.069          & 0.050           & 0.034          & 0.155          & \textbf{0.135} \\
\hspace{2mm}Scale Agnostic & 0.073          & \underline{0.033} & \underline{0.027} & 0.067          & \underline{0.029} & \underline{0.148} & \textbf{0.136} \\
\midrule
\textbf{Ours} & \textbf{0.035} & 0.118          & 0.068          & \textbf{0.038} & 0.041          & \textbf{0.025} & \underline{0.143 }         \\
\bottomrule
\end{tabular}}
\label{tab: Ablation Easy}
\end{table}

\subsection{Datasets and Implementation}

\paragraph{TartanAir dataset}
The Tartanair dataset is a large-scale synthetic dataset encompassing highly diverse scenes, including various complex and challenging environments. 
Following the TartanAir data generation method, we created new, more diverse and challenging trajectories. 
From these, we selected that feature fast camera movements, low-light indoor environments, and simulated lunar surfaces lacking visual features. The images in the new dataset have a resolution of \(640 \times 640\) . 
When testing iSLAM-VO and TartanVO, we resized the input images to \(448 \times 640\) to match the input requirements of the optical flow network. Due to the substantial GPU memory required by DROID-SLAM for global bundle adjustment optimization when processing high-resolution images, we reduced the input image resolution to \(512 \times 512 \) and manually reclaimed GPU and memory after testing each trajectory to avoid potential memory insufficiency.

\paragraph{KITTI dataset}
The KITTI dataset is a well-known and widely used dataset for autonomous driving, containing detailed ground truth labels that make it suitable for evaluating the performance of various VO/SLAM methods. 
Learning-based methods may experience performance degradation when handling the image sizes in the KITTI dataset, therefore, we cropped the input images to different sizes based on the methods tested. Our model processed images cropped to \( 376\times680\) , while for testing DROID-SLAM, the images were cropped to \( 320 \times 832\) .

\paragraph{EuRoC dataset}
 The Euroc dataset consists of 11 trajectories collected by a drone and is also a widely used benchmark for VO/SLAM tasks. Some scenes in this dataset contain thousands of image pairs, which imposes computational pressure on DROID-SLAM when performing bundle adjustment. Thus, during testing, we reduced the input image resolution to \( 320 \times 512\) and read every other frame of image pairs. In post-processing, we completed the entire trajectory by interpolating timestamps.

\paragraph{ZED Camera data}
The ZED Camera data comprises real-world trajectory data captured using a ZED stereo camera, employed to test the robustness and generality of our model when faced with unseen data not present in the training set.
% \section{Network Details}

\subsection{Network and Training Details}
\label{appendix:Network}
% In the uncertainty-aware matching network, we use a decoder for global motion features similar to the Flowformer to obtain the $\Delta \sigma$ after each iteration. 
In the uncertainty-aware matching network, we employ a flow decoder similar to Flowformer to obtain $\Delta f$. 
In each iteration, the flow $f_x$ is updated by $f_x \leftarrow f_x + \delta f$.
With the shared features, we finetune a new decoder with \texttt{ConvGRU} layers to estimate the uncertainty updates $\Delta \sigma$.
In each iteration, the matching uncertainty is updated by $\sigma \leftarrow \sigma \times \Delta \sigma$.

For each pixel $x$, we extract the local cost-map patch $q_x$ from the cost map, and the context feature $t_x$ from the context encoder.
We encode the $q_x$ using a transformer $\text{FFN}$.
Given the 2D position $p = x + f_x $, we encode it into positional embedding $\text{PE}(p)$.
We aggregate these features using CNN encoder $\text{ME}$ to generate local motion features ${I_x}$.
Utilizing the GMA module, the network's global motion features $G_x$ are acquired from the current motion features $\text{Att}_f(t_x)$ and ${I_x}$. 
Subsequently, the shared motion features $m_x$ is obtained by concatenating the $G_x$, $I_x$ and $\text{Att}(t_x)$, where $\text{Att}(t_x)$ is the attention of the context features.
To estimate the flow update $\Delta f$ and $\Delta \Sigma$, we use the ConvGRU $\Sigma_\pi (m_x)$
% The FlowGRU and CovGRU utilize different motion encoders $\text{Att}$, respectively, and the GRU modules leverage the same global motion features to separately obtain $\Delta f$ and $\Delta \sigma$.
% the encoded features $\text{FFN}(q_x)$ are first obtained, which are encoded by the local cost-map patch $q_x$. 
% These features are then combined with the positional embedding $\text{PE}(p)$ where $p$ is matched point $p = x + f_x $
% using the Motion Encoder $\text{ME}$ to derive the intermediate tensor ${inp}$. 
% Here, $t_x$ denotes the encoded context features. 

\begin{equation}
\begin{aligned}
        I_x &= \text{ME}(f_x, \text{Concat}(\text{FFN}(q_x), \text{PE}(p))) \\
    G_x&=\text{GMA}(\text{Att}(t_x), {I_x}) \\
    m_x &= \text{Concat}(\text{Att}(t_x), {I_x}, G_x) \\
    \Delta f&=\text{FlowGRU}(m_x) \\
    \Delta \Sigma&=\text{CovGRU}(m_x) \\
\end{aligned}
\end{equation}

% \subsection{Training details}

We use \texttt{AdamW} optimizer with a learning rate of $12.5 \times 10^{-5}$. The model is trained on A100 GPU consuming 16 GB GPU memory.

\subsection{GPU memory \& Parameters}

% In this section, we show the GPU memory and parameters we used during testing.
% why our  consume less GPU memory
As shown in \tref{table:memory}, we use 4.20GB of GPU memory, which is 6.7 times smaller than DROID-SLAM.
This is because we utilize the sparse features for back-end optimization, reducing the requirements for the GPU memory.

% Our model occupies 3.2GB of GPU memory when run on the EuRoC dataset at the original image scale.

% Methods & GPU memory & Parameters 
% Droid SLAM
% DPVO
% Ours (Small) 
% Ours

\begin{table}[t]
\centering
\caption{The GPU memory consumption and the number of parameters during testing}
\vspace{2mm}
\resizebox{.6\linewidth}{!}{\begin{tabular}{lcc}
\toprule
  \textbf{Methods} & \textbf{GPU memory} & \textbf{Parameters} \\
  \midrule
  \textbf{Ours} & 4.20GB & 19.2M \\
  \midrule
  TartanVO-stereo & 1.16GB & 47.3M \\
  DROID-SLAM & 28.34GB & 4.0M \\
  DPVO & 1.51GB & 3.4M \\
  \bottomrule
\end{tabular}}
\label{table:memory}
\end{table}

\section{Runtime Analysis}
\label{appendix:RuntimeAnalysis}

\begin{figure}[t]
    \centering
    \includegraphics[width=\linewidth]{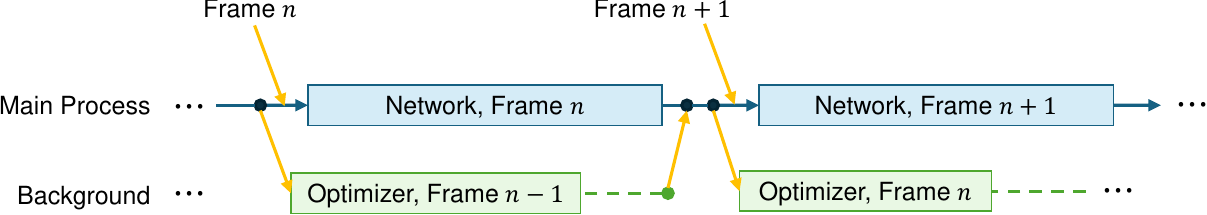}
    \caption{Multiprocessing setup places the optimizer in the background process and parallelizes the CPU-intensive (optimizer) and GPU-intensive (frontend network inference) jobs.}
    \label{fig: multiprocesspipeline}
\end{figure}

The runtime analysis use the platform with \texttt{AMD Ryzen 9 5950X} CPU and \texttt{NVIDIA 3090 Ti} GPU.
% The sequence data is loaded into RAM before timing starts to reduce the overhead caused by disk I/O.
To speed up the frontend network, we utilize  TensorRT\footnote{\href{https://developer.nvidia.com/tensorrt}{https://developer.nvidia.com/tensorrt}} (\textbf{TRT}), and multi-processing (\textbf{MP}) to accelerate the original network (\textbf{Raw}).
In our multiprocessing setup, as shown in \fref{fig: multiprocesspipeline}, a new CPU process is initiated to run the pose graph optimization in parallel with the front-end network, thereby maximizing GPU utilization.

We also introduce a fast mode (\textbf{MAC-VO Fast}) which utilizes half-precision (\texttt{bfloat16}) number in the network inference to enhance computational efficiency.
This mode also speeds up the memory decoder network by reducing the number of iterative updates from 12 to 4. 
The fast mode performs \textbf{10.5 fps} (frames per second) with 70\% of the performance of the original MAC-VO.

\end{document}